\documentclass[10pt,twocolumn,letterpaper]{article}

\usepackage{cvpr}      

\usepackage{graphicx}
\usepackage{amsmath}
\usepackage{amssymb}
\usepackage{booktabs}
\usepackage{multirow}

\usepackage[pagebackref,breaklinks,colorlinks]{hyperref}

\usepackage[capitalize]{cleveref}
\crefname{section}{Sec.}{Secs.}
\Crefname{section}{Section}{Sections}
\Crefname{table}{Table}{Tables}
\crefname{table}{Tab.}{Tabs.}

\def\cvprPaperID{*****} 
\def\confName{CVPR}
\def\confYear{2023}

\begin{document}

\title{Foreground Object Search by Distilling Composite Image Feature}

\author{Bo Zhang$^{1}$ \and
Jiacheng Sui$^{2}$ \and
Li Niu\thanks{Corresponding author} $^{1}$ \and
$^1$ Center for Machine Cognitive Computing of Artificial Intelligence Institute \\ Artificial Intelligence Institute, Shanghai Jiao Tong University \\
{\tt\small\{bo-zhang, ustcnewly\}@sjtu.edu.cn} \and
$^2$ Xian Jiao Tong University \\
{\tt\small rookiecharles99@gmail.com}
}

\maketitle

\begin{abstract}
   Foreground object search (FOS) aims to find compatible foreground objects for a given background image, producing realistic composite image. We observe that competitive retrieval performance could be achieved by using a discriminator to predict the compatibility of composite image, but this approach has unaffordable time cost. To this end, we propose a novel FOS method via \textbf{dis}tilling \textbf{co}mposite feature (DiscoFOS). Specifically, the abovementioned discriminator serves as teacher network. The student network employs two encoders to extract foreground feature and background feature. Their interaction output is enforced to match the composite image feature from the teacher network. Additionally, previous works did not release their datasets, so we contribute two datasets for FOS task: S-FOSD dataset with synthetic composite images and R-FOSD dataset with real composite images. Extensive experiments on our two datasets demonstrate the superiority of the proposed method over previous approaches. The dataset and code are available at \url{https://github.com/bcmi/Foreground-Object-Search-Dataset-FOSD}.
\end{abstract}

\section{Introduction}
\label{sec:intro} 
\begin{figure}[t]
    \begin{center}
    \includegraphics[width=1\linewidth]{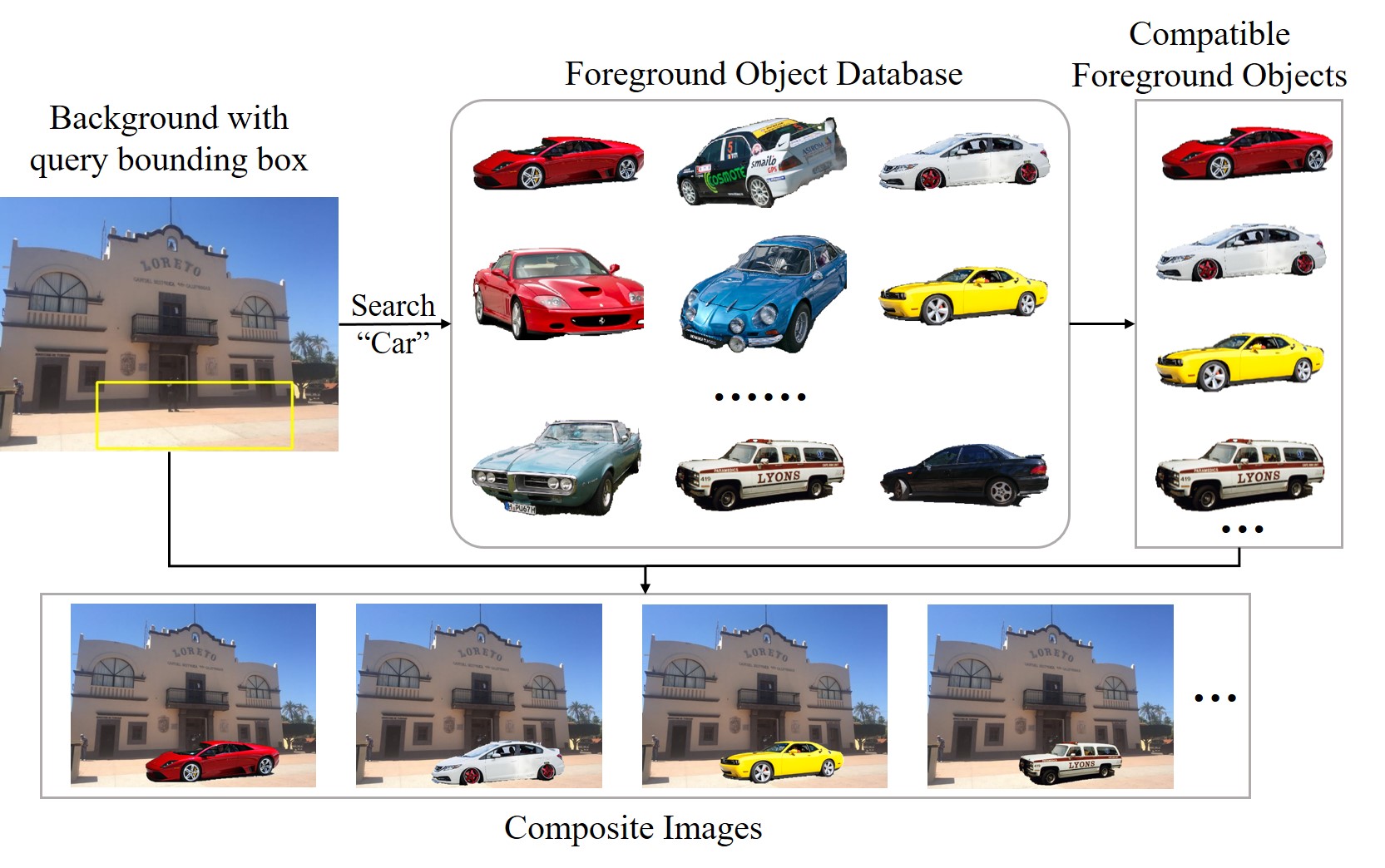}
    \end{center}
    \caption{Illustration of foreground object search. Given a background image with query bounding box (yellow), foreground object search aims to find compatible foreground objects of a specified category from a database, which is composited with the background to produce a realistic composite image.}
    \label{fig:fos_illustration}
\end{figure}

Foreground Object Search (FOS) aims to find compatible foregrounds from specified category for a given background image which has a query bounding box indicating the foreground location \cite{Zhao2018CompositingAwareIS}. More precisely, an object is compatible with a background image if it can be realistically composited into the image \cite{Zhao2019UnconstrainedFO}, as illustrated in Figure~\ref{fig:fos_illustration}. FOS is a core technique in many image composition applications \cite{Niu2021MakingIR}. For example, FOS technique can help users acquire suitable foregrounds from a foreground pool automatically and efficiently for object insertion in photo editing \cite{Li2020InterpretableFO}. Moreover, FOS also can be used to fill a region comprising undesired objects using new foreground \cite{Zhao2019UnconstrainedFO}. 

There exist many factors that affect the compatibility between background and foreground, including semantics, style (\eg, color and texture), lighting, and geometry (\eg, shape and viewpoint). Previous works on FOS may consider different factors. For example, early methods \cite{Zhao2018CompositingAwareIS,Zhao2019UnconstrainedFO} focused on the semantic compatibility. Recent works~\cite{Li2020InterpretableFO, Wu2021FinegrainedFR, Zhu2022GALATG} considered geometry and other factors, including style~\cite{Wu2021FinegrainedFR} and lighting~\cite{Zhu2022GALATG}.
In this paper, we focus on semantics and geometry compatibility following~\cite{Li2020InterpretableFO}, because incompatible color and lighting between the background and foreground can be tackled to some extent by image harmonization~\cite{CongDoveNet2020,Ling2021RegionawareAI,Cong2021BargainNetBD}.

The general pipeline of most existing methods \cite{Zhao2018CompositingAwareIS,Zhao2019UnconstrainedFO,Zhu2022GALATG,Li2020InterpretableFO, Wu2021FinegrainedFR} is to learn an embedding space with two encoders respectively for background and foreground, so that compatible background and foreground are close to each other in this space. Alternatively, some approaches~\cite{Zhu2015LearningAD, Zhao2019UnconstrainedFO} trained a discriminator to predict background-foreground compatibility by feeding the composite image. Based on preliminary experiments, we observe that the discriminator can achieve much better results than encoders when taking the cropped composite image as input (see teacher network in Figure~\ref{fig:framework}).
We conjecture that the forward pass in the discriminator allows thorough interaction between background and foreground, which could provide useful contextual cues for estimating background-foreground compatibility. However, given a background image, it is very time-consuming to composite with each foreground image and perform forward computation for each composite image. Motivated by this, we propose a novel FOS framework called DiscoFOS via knowledge distillation.  
Specifically, we distill the knowledge of composite image from the discriminator to two encoders, in which we enforce the interaction output of foreground feature and background feature to match with the composite image feature. How to design the interaction between two encoders is challenging, due to the trade-off between performance and computational cost. On the one hand, insufficient interaction between two encoders may be unable to mimic the rich knowledge in composite image feature. On the other hand, sufficient interaction would largely increase the computational burden. Considering the abovementioned trade-off, we perform interaction only on the last feature maps of two encoders, which achieves significant performance improvement with acceptable computational overhead.

Since previous works (CAIS \cite{Zhao2018CompositingAwareIS}, UFO \cite{Zhao2019UnconstrainedFO}, and GALA \cite{Zhu2022GALATG}) did not release their datasets, we build our own datasets based on an existing large-scale real-world dataset, \ie, Open Images \cite{Kuznetsova2020TheOI}, as illustrated in Figure~\ref{fig:dataset_construct}. We construct two FOS Datasets respectively containing Synthetic composite images and Real composite images, abbreviated as S-FOSD and R-FOSD respectively. 
We first introduce the S-FOSD dataset. Given a real image with instance segmentation mask, we choose one object and fill its bounding box with image mean values to get the background. Meanwhile, we crop out this object as foreground. After removing unsuitable categories and occluded foregrounds, the resultant dataset contains 57,859 backgrounds and 63,619 foregrounds. Following \cite{Zhao2018CompositingAwareIS,Zhu2022GALATG}, for each background image, we deem the foreground object from the same image as ground-truth. 
For R-FOSD dataset, we collect images from Internet as background images and draw a bounding box at the expected foreground location as query bounding box. R-FOSD dataset shares the same foregrounds with the test set of S-FOSD dataset. Then we employ multiple human annotators to label the compatibility of each pair of background and foreground.
\emph{In summary, S-FOSD dataset is lowcost and highly scalable, but has neither complete background nor ground-truth negative samples. Oppositely, R-FOSD dataset has complete background image with both positive and negative foregrounds annotated by human, yet is unscalable due to the high annotation cost.} In our experiments, S-FOSD dataset is used for both training and validating model, while R-FOSD dataset is only used for model evaluation. More details about dataset construction could be found in Section \ref{sec:dataset_construct}.

We evaluate our method on the proposed datasets, which validates the superiority of the proposed method over previous approaches. Our major contributions can be summarized as follows:
1) To facilitate the research on FOS task, we contribute two public datasets: S-FOSD dataset with synthetic composite images and R-FOSD dataset with real composite images.
2) We propose a novel method named DiscoFOS that improves foreground object search by distilling the knowledge of composite image feature into two encoders.
3) Extensive experiments demonstrate the superiority of the proposed method over previous baselines on our datasets.

\section{Related Works}
\label{sec:related_works}

\subsection{Foreground Object Search}
Early works \cite{Lalonde2007PhotoCA,Chen2009Sketch2PhotoII} employed hand-crafted features to match background with foreground, yet their performance may be limited by the representation ability of hand-crafted features. Recent work applied deep learning based feature for foreground retrieval. For example, \cite{Tan2018WhereAW} utilized deep features to capture local context particularly for person compositing. 
\cite{Zhu2015LearningAD} trained a discriminator to estimate the realism of composites, which is available for selecting compatible foregrounds by compositing each foreground with the background, but is computationally expensive. 

More recent methods~\cite{Zhao2018CompositingAwareIS,Zhao2019UnconstrainedFO,Zhu2022GALATG,Wu2021FinegrainedFR,Li2020InterpretableFO} typically trained two encoders to extract background feature and foreground feature, and then measured background-foreground compatibility by calculating feature similarity. These methods considered different factors that affect the compatibility between background and foreground. For example, the methods~\cite{Zhao2018CompositingAwareIS,Zhao2019UnconstrainedFO} considered the semantic compatibility, while the approaches~\cite{Li2020InterpretableFO, Wu2021FinegrainedFR, Zhu2022GALATG} considered geometry and other factors, including style~\cite{Wu2021FinegrainedFR} and lighting~\cite{Zhu2022GALATG}. In this work, we focus on semantic compatibility and geometry compatibility following~\cite{Li2020InterpretableFO}, and propose a novel foreground object search method that improves the encoders with knowledge distillation. Additionally, previous works \cite{Zhao2018CompositingAwareIS,Zhao2019UnconstrainedFO,Wu2021FinegrainedFR,Zhu2022GALATG} did not release their datasets, we contribute two datasets to facilitate the research in this field.

\subsection{Image Composition}
As summarized in~\cite{Niu2021MakingIR}, existing image composition works attempted to solve one or some issues which affect the quality of composite image,  such as illumination, shadow, and geometry.
Image harmonization methods \cite{TsaiDIHarmonization2017, CongDoveNet2020, IHCISSAMCun2020,cong2022high} focused on eliminating the color and illumination discrepancy between background and foreground.
Besides, some shadow generation works \cite{ExposingKee2014, DPILEGardner2019, AllWeatherZhang2019, ARShadowGANLiu2020, hong2021shadow} aimed to generate plausible shadow for the inserted foreground.
To blend the foreground into the background more naturally, image blending methods \cite{PoissoneditingPatrick2003, ClosedMattingLevin2006, GPGANWu2019} paid attention to smoothing the boundary between background and foreground. 
The closest subtask to ours is object placement~\cite{Zhou2022LearningOP, Niu2022FastOP, Liu2021OPAOP, ContextawareLee2018, SyntheticTripathi2019}, which generates reasonable locations and sizes to place foreground over background. Given a background, object placement focuses on predicting bounding box for the inserted foreground, while our task targets at retrieving compatible foregrounds from a specified category for a given query bounding box. 

\subsection{Knowledge Distillation}
Knowledge distillation methods usually improve the student network by forcing to mimic the behaviors of the teacher network, such as soft predictions \cite{Hinton2015DistillingTK}, logits \cite{Ba2014DoDN}, intermediate feature maps \cite{Romero2015FitNetsHF}, or attention maps \cite{Zagoruyko2017PayingMA}. 
Apart from the above works on image classification, some recent works extend knowledge distillation to more complex vision tasks, including semantic segmentation \cite{Zhang2022DistillingID, Shu2021ChannelwiseKD}, object detection \cite{Li2017MimickingVE, Chen2017LearningEO}, face recognition \cite{Huang2022EvaluationorientedKD}, and so on. 

Previous works \cite{Li2020InterpretableFO, Wu2021FinegrainedFR} have also considered introducing knowledge distillation to foreground object search, in which foreground (\emph{resp.}, background) encoder plays the role of the teacher (\emph{resp.}, student) network and the foreground information is distilled from foreground embedding to background embedding. However, these approaches rely on fine-grained annotations of foreground attributes~\cite{Li2020InterpretableFO} or multiple pretrained models~\cite{Wu2021FinegrainedFR}, which may hinder the generalization to unseen foreground categories.  
Unlike the above methods, our method adopts a composite image discriminator as teacher network and two encoders as student network, in which we distill composite image feature from discriminator to the interaction output of foreground feature and background feature, which proves to be effective and computationally affordable.

\begin{figure}[t]
    \begin{center}
    \includegraphics[width=1\linewidth]{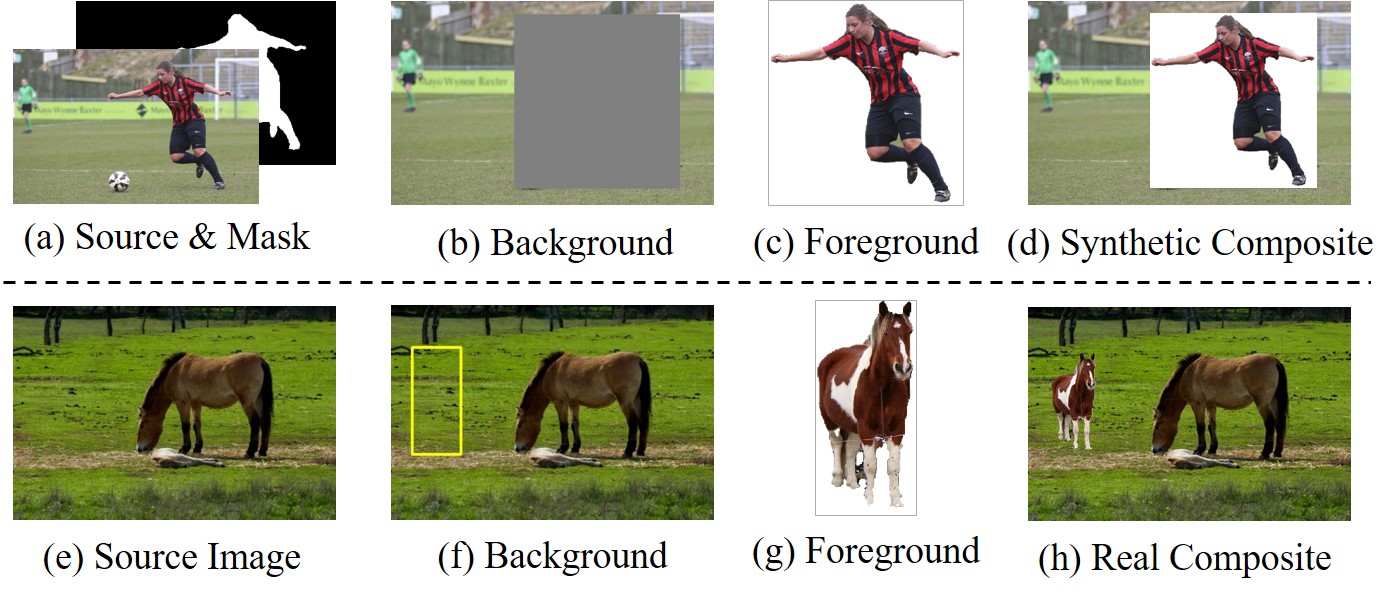}
    \end{center}
    \caption{The illustration of building our foreground object search (FOS) datasets. Top: FOS Dataset with Synthetic composite images (S-FOSD). Bottom: FOS Dataset with Real composite images (R-FOSD). More details about dataset construction can be seen in Section~\ref{sec:dataset_construct}. 
    }
    \label{fig:dataset_construct}
\end{figure}

\section{Dataset Construction}
\label{sec:dataset_construct}
Our datasets are constructed on existing Open Images dataset \cite{Kuznetsova2020TheOI} that contains 9.2 million images covering diverse scenes, making it very suitable for real-world evaluation.
We build two datasets for foreground object search: S-FOSD with synthetic composite images and R-FOSD with real composite images, which have different ways to acquire backgrounds and foregrounds (see Figure~\ref{fig:dataset_construct}).

\subsection{S-FOSD Dataset}
\label{sec:synthetic_dataset}

\noindent\textbf{Foreground Object Selection.}  Open Images dataset \cite{Kuznetsova2020TheOI} provides instance segmentation masks for 2.8 million object instances in 350 categories. To accommodate our task, we delete the categories and objects that are unsuitable for the task or beyond our focus (geometry and semantic compatibility), after which 32 categories remain.  
The detailed rules for object selection and the complete category list can be found in Supplementary.

\noindent\textbf{Background and Foreground Generation.} By using the segmentation masks, we generate background and foreground images in a similar way to previous works \cite{Zhao2018CompositingAwareIS,Zhu2022GALATG,Zhao2019UnconstrainedFO}. Specifically, as shown in Figure~\ref{fig:dataset_construct} (a) $\sim$ (c), given an image with instance segmentation masks, we choose one object and fill its bounding box with image mean values to get the background. Meanwhile, we crop out the object and paste it on white background, producing the foreground. With the background and foreground, we can obtain the synthetic composite image by resizing the foreground and placing it in the query bounding box on the background (see Figure~\ref{fig:dataset_construct} (d)).  
Additionally, we have also tried image inpainting~\cite{suvorov2021resolution,zhao2021comodgan, yu2018generative} to fill the object bounding box, but got unsatisfactory results, probably due to large missing content and the residues of erased object (\eg, shadow). 
After that, we obtain over 63,000 pairs of background and foreground covering 32 categories, with a maximum of 5,000 images and a minimum of 500 images in each category. For a given background, the foreground cropped from the same image is naturally compatible and thus deemed as ground-truth following \cite{Zhao2018CompositingAwareIS,Zhu2022GALATG,Zhao2019UnconstrainedFO}.

\noindent\textbf{Dataset Split.} S-FOSD dataset is employed for both training and testing, as its construction is low-cost and highly scalable. We first select test samples to build the test set and the rest forms the training set. For reliable performance evaluation, we build test set mainly concerning its diversity and quality. Finally, we get 20 backgrounds and 200 foregrounds for each category, which contains 20 pairs of background and foreground. The remaining 57,219 pairs of background and foreground form the training set.

\begin{figure*}[t]
    \begin{center}
    \includegraphics[width=0.9\linewidth]{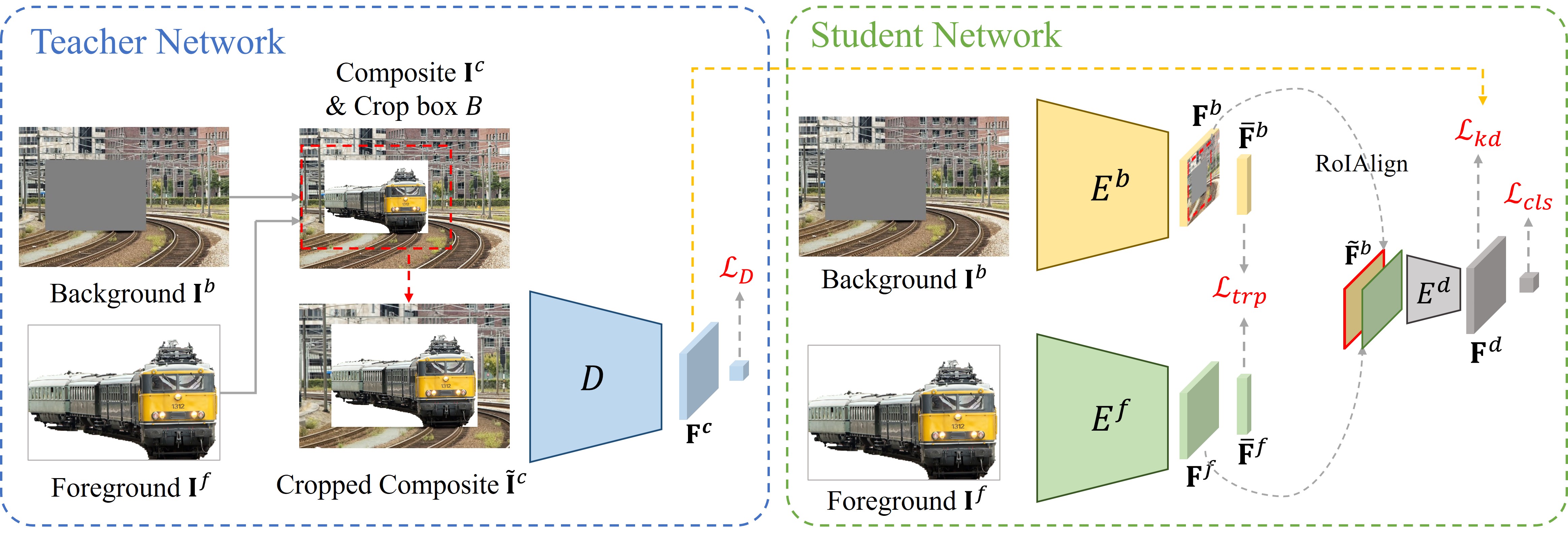}
    \end{center}
    \caption{Illustration of the proposed DiscoFOS for foreground object search.
    The discriminator $D$ is first trained to predict the background-foreground compatibility of input composite image $\mathbf{I}^c$, whose intermediate feature map $\mathbf{F}^c$ then servers as distillation target to train the student model. The student network first extracts background feature $\mathbf{F}^b$ and foreground feature $\mathbf{F}^f$ by two encoders $\{E^f,E^b\}$, then applies the features to generate distilled feature $\mathbf{F}^d$ and compatibility prediction by knowledge distillation module $E^d$.}
    \label{fig:framework}
\end{figure*}

\subsection{R-FOSD Dataset}
\label{sec:real_dataset}

\noindent\textbf{Background and Foreground Generation.} When building R-FOSD dataset, we directly adopt the foregrounds of the test set in S-FOSD dataset and collect images from Internet as backgrounds. It is very likely that a random image is unsuitable for compositing with any test foreground. Thus, we collect candidate backgrounds by searching similar images to the test backgrounds of S-FOSD dataset. After that, we draw a bounding box at the desired foreground location as query bounding box (see Figure~\ref{fig:dataset_construct} (f)). The size and location of the bounding box are decided by first mimicking the foreground in the similar background of S-FOSD dataset and then manual inspection. 
For each pair of background and foreground, we resize the foreground and place it in the query bounding box on the source image, generating a real composite image (see Figure~\ref{fig:dataset_construct} (h)).
Finally, R-FOSD dataset contains 32 categories, each of which has 20 background images and 200 foreground images.

\noindent\textbf{Compatibility Labelling.} To acquire the binary compatibility label (1 for compatible background-foreground pair and 0 for incompatible pair), we employ three human annotators to label the compatibility for $32 \times 20 \times 200$ pairs of background and foreground. During annotation, we show real composite images and request annotators to assign binary labels by considering the semantics and geometry compatibility between background and foreground. Finally, for one background, we only consider the foregrounds for which all three human annotators label 1 as compatible and the others are treated as incompatible. Note that manually annotating the dataset is expensive, as it requires labelling a quadratically growing number of background and foreground pairs. Therefore, we only use R-FOSD dataset as a test set. 
To keep consistent with training data, we also fill the query bounding box of the background in R-FOSD dataset with image mean values at test time and the complete background image of R-FOSD dataset is only used to obtain the final composite image (Figure~\ref{fig:dataset_construct} (h)).

\section{Methodology}
\label{sec:methodology}
In this section, we describe our proposed method for foreground object search. As illustrated in Figure~\ref{fig:framework}, we train a discriminator $D$ to predict the compatibility of composite image, which serves as the teacher network (see Section~\ref{sec:comp_discriminator}). We employ two encoders $E^b$ and $E^f$ (see Section~\ref{sec:fg_bg_encoders}) as well as a light-weight knowledge distillation module $E^d$ (see Section~\ref{sec:kd_network}) as the student network. The two encoders respectively extract background feature $\mathbf{F}^b$ and foreground feature $\mathbf{F}^f$, which are fed into distillation module to interact with each other. During training stage, we enforce the interaction output $\mathbf{F}^d$ to match with the composite image feature $\mathbf{F}^c$, in which the background-foreground compatibility information is distilled to $\mathbf{F}^d$. Finally, we predict compatibility scores for pairwise background and foreground based on the distilled feature $\mathbf{F}^d$.        

\subsection{Composite Image Discriminator}
\label{sec:comp_discriminator} 
\noindent\textbf{Network Architecture.} Given a background image $\mathbf{I}^b$ with a query bounding box and a foreground image $\mathbf{I}^f$, we first generate a synthetic composite image $\mathbf{I}^c$ as in Figure~\ref{fig:dataset_construct} (d). Then, we employ a discriminator $D$ that takes a composite image as input to predict whether background and foreground of the composite image are compatible. We implement the discriminator as a binary classifier that consists of backbone network and classification head. Moreover, based on preliminary experiments, we find that feeding a cropped composite image into the discriminator achieves significantly better performance than feeding the whole composite image. Specifically, we crop the composite image, ensuring that the foreground object is located at the center and the area of foreground bounding box is about 50 percent of the whole crop.
Then, we resize the crop to the input size 224 $\times$ 224, which is denoted as $\mathbf{\widetilde{I}}^c$. The superiority of feeding cropped composite image can be attributed to that the contextual information near the foreground may be more helpful and the foreground is aligned with crop center. Here we refer to the crop box as $B$. 
With the cropped composite image, the discriminator first extracts the composite image feature map $\mathbf{F}^c$ by backbone network and then apply global average pooling (GAP) layer followed by a binary classifier. We train the discriminator using binary cross-entropy loss:
\begin{equation}
    \mathcal{L}_{D} = -\log \left( p(\mathbf{\widetilde{I}}^c)_{y} \right),
\label{eqn:dis_loss}
\end{equation}
in which $p(\cdot)_y$ means the predicted probability corresponding to the ground-truth compatibility label $y$.    

\begin{figure}[t]
    \begin{center}
    \includegraphics[width=1\linewidth]{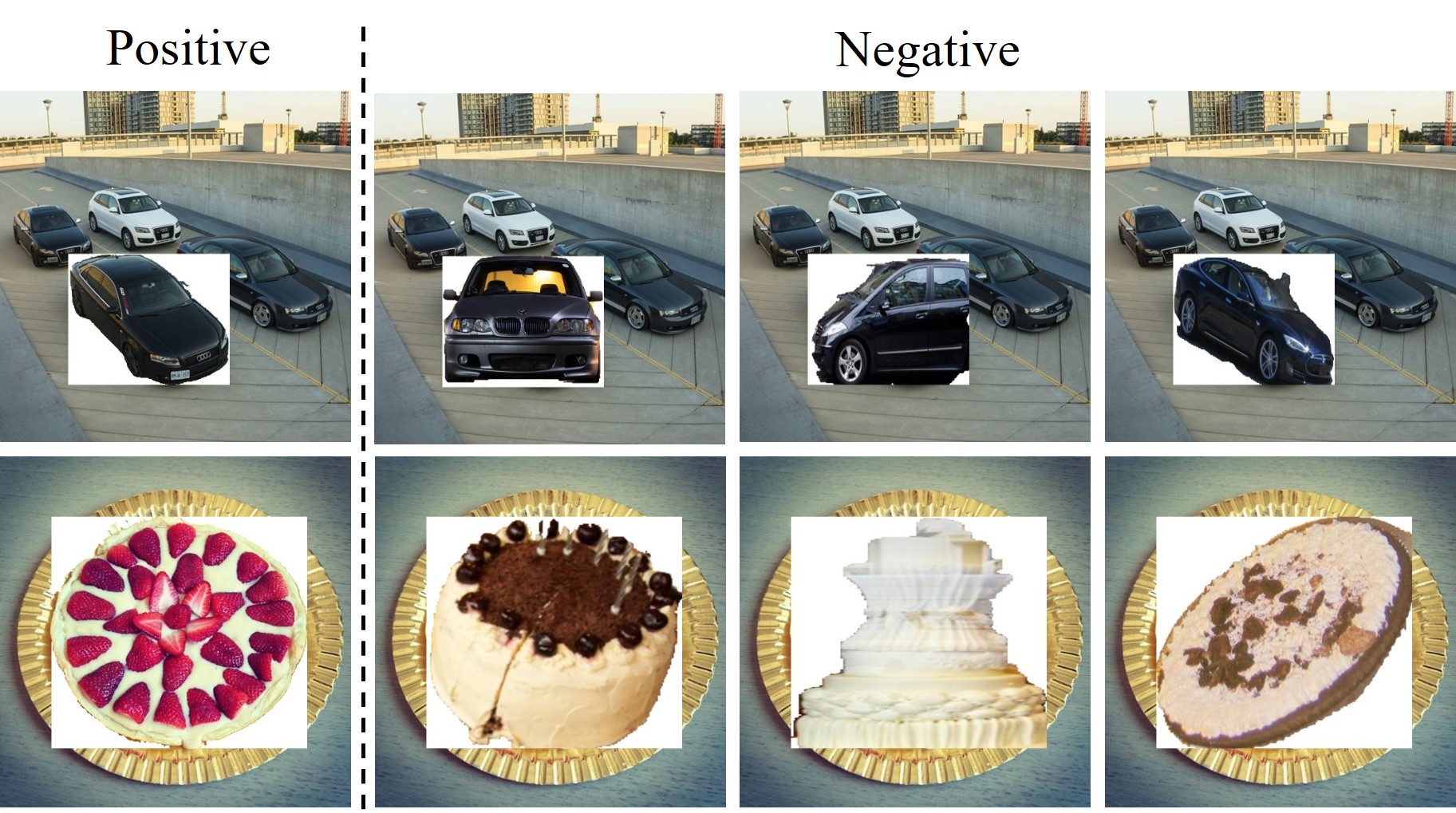}
    \end{center}
    \caption{Examples of positive and negative samples used to train our models. Given a background in S-FOSD dataset, we have one positive foreground and one or more negative foregrounds, which are filled with white background pixels.
    }
    \label{fig:train_examples}
\end{figure}

\noindent\textbf{Positive and Negative Samples Generation.} During training phase, we consider compatible (\emph{resp.}, incompatible) foreground objects as positive (\emph{resp.}, negative) samples for a given background. The foreground cropped from the same image as the background is naturally viewed as positive sample. However, other foreground objects may also be compatible with the background. To guarantee the effectiveness of training samples, similar to \cite{Zhao2019UnconstrainedFO}, we train a binary classifier using VGG-19 \cite{Simonyan2015VGG} pretrained on ImageNet~\cite{deng2009imagenet} as backbone network to help filter out training samples. Given a pair of background image and foreground image, the classifier takes their composite image as input and predicts the compatibility score. When training this classifier, we assume that only background-foreground pairs from same images are positive and others are negative. With the trained classifier, we restrict negative samples to only include those foreground objects which are confidently classified as incompatible, that is, compatibility score is lower than a threshold (0.3 in our experiments). Therefore, given a background image, we have a single positive foreground and one or more negative foregrounds in S-FOSD dataset (see Figure~\ref{fig:train_examples}). The obtained positive and negative samples are used to train the composite image discriminator in this section and the student network in Section~\ref{sec:fg_bg_encoders}, \ref{sec:kd_network}.

\subsection{Background and Foreground Encoders}
\label{sec:fg_bg_encoders}
We employ two encoders $E^b$ and $E^f$ to extract feature respectively from background image $\mathbf{I}^b$ and foreground image $\mathbf{I}^f$. Correspondingly, the background encoder $E^b$ outputs background feature map $\mathbf{F}^b$ and the foreground encoder $E^f$ outputs foreground feature map $\mathbf{F}^f$, both of which have the same shape, \ie, $\mathbf{F}^b, \mathbf{F}^f \in \mathcal{R}^{h \times w \times c}$. Here $h \times w$ is the spatial size and $c$ is the number of channels. Then we apply GAP layer to yield background and foreground feature vectors, denoted as $\mathbf{\bar{F}}^b, \mathbf{\bar{F}}^f \in \mathcal{R}^{c}$. After mapping the foreground and background into the common feature space, we enforce the compatible background and foreground to be close to each other in this space. Thus, the compatibility between foreground and background can be measured by computing the cosine similarity between their features. To this end, following previous works~\cite{Wu2021FinegrainedFR,Zhao2018CompositingAwareIS,Zhao2019UnconstrainedFO,Zhu2022GALATG}, we train the encoders using triplet loss \cite{Schroff2015FaceNetAU}, which tends to pull the positive sample (\ie, compatible foreground) to the anchor (\ie, given background) and push the negative sample (\ie, incompatible foreground) away from the anchor in the feature space. During training stage, we tend to minimize the following loss function:
\begin{equation}
    \mathcal{L}_{trp} =\max \left(0, m + S(\mathbf{\bar{F}}^b, \mathbf{\bar{F}}^f_n) - S(\mathbf{\bar{F}}^b, \mathbf{\bar{F}}^f_p)  \right),
\label{eqn:triplet_loss}
\end{equation}
where $S(\cdot)$ represents cosine similarity and $m$ is a positive margin in the range of (0, 1). $\mathbf{\bar{F}}^f_p$ (\emph{resp.}, $\mathbf{\bar{F}}^f_n$) is the feature of positive (\emph{resp.}, negative) foreground for background $\mathbf{\bar{F}}^b$. By minimizing $\mathcal{L}_{trp}$, for a given background, the feature similarity with compatible foreground is expected to be greater than that with incompatible foreground by the margin $m$.
Additionally, the above encoders are usually employed in previous methods~\cite{Zhao2018CompositingAwareIS,Zhao2019UnconstrainedFO,Zhu2022GALATG}. Given a query background and a foreground database, these approaches rank the compatibility of different foregrounds by measuring their feature similarity to the given background.

\subsection{Knowledge Distillation Module}
\label{sec:kd_network}
After evaluating the discriminator in Section~\ref{sec:comp_discriminator} and the encoders in Section~\ref{sec:fg_bg_encoders}, we observe that discriminator can achieve much better results than encoders, probably because the forward propagation in the discriminator allows thorough interaction between background and foreground, providing useful contextual cues for estimating background-foreground compatibility.
However, retrieving foreground images using discriminator requires to composite with each foreground image, which brings heavy computational burden. On the contrary, encoders achieve inferior performance, yet have significantly faster speed due to the exemption from background-foreground interaction. This motivates us to distill discriminator knowledge into encoders.

To achieve this goal, we design a knowledge distillation module $E^d$, which interacts foreground feature with background feature and enforces the interaction output to match with the composite image feature $\mathbf{F}^c$ from the discriminator. Recall that the composite image feature is extracted from cropped composite image, with the crop bounding box denoted as $B$ (see Section~\ref{sec:comp_discriminator}). So we apply RoIAlign~\cite{He2020MaskR} with the bounding box $B$ to obtain local background feature map $\mathbf{\widetilde{F}}^b$ from global background feature map $\mathbf{F}^b$ produced by background encoder $E^b$. Then, we resize RoIAlign output to be of the same shape as $\mathbf{F}^b$, \ie, $\mathbf{\widetilde{F}}^b \in \mathcal{R}^{h \times w \times c}$. The local background feature is supposed to encode contextual information surrounding the foreground. Meanwhile, we utilize foreground encoder $E^f$ to extract foreground feature map $\mathbf{F}^f$. Given the foreground feature map $\mathbf{F}^f$ and local background feature map $\mathbf{\widetilde{F}}^b$, we feed their concatenation into $E^d$ to produce a distilled feature map $\mathbf{F}^d \in \mathcal{R}^{h \times w \times c}$. Note the discriminator and two encoders adopt the same backbone network and input image size in our implementation, so their feature maps have the same shape, \ie, $\mathbf{F}^c, \mathbf{F}^d \in \mathcal{R}^{h \times w \times c}$. During training phase, we enforce the distilled feature $\mathbf{F}^d$ to mimic the composite image feature $\mathbf{F}^c$ by $L_1$ loss:
\begin{equation}
    \mathcal{L}_{kd} = \|\mathbf{F}^d- \mathbf{F}^c \|_1.
\label{eq:kd_loss}
\end{equation}
We pool the distilled feature map into a vector and send it to a binary classifier to predict the compatibility. The classifier is also trained using binary cross-entropy loss:
\begin{equation}
    \mathcal{L}_{cls} = -\log \left(p(\mathbf{F}^d)_{y} \right),
\label{eqn:cls_loss}
\end{equation}
in which $p(\cdot)_y$ is similarly defined as in Eqn. \ref{eqn:dis_loss}. 

Finally, we train the encoders $\{E^f,E^b\}$ and distillation module $E^d$ simultaneously. The overall optimization function can be written as
\begin{equation}
\label{eqn:total_loss}
    \mathcal{L} = \mathcal{L}_{trp} + \lambda_{kd} \mathcal{L}_{kd} + \lambda_{cls} \mathcal{L}_{cls},
\end{equation}
where $\lambda_{kd}$ and $\lambda_{cls}$ are trade-off parameters.
During inference, our model finds compatible foregrounds for a given background by ranking the predicted compatibility scores.

\subsection{Generalization to Real-world Application}
\label{sec:generalization_realworld}
To boost the performance of our method on real-world data, we make some modifications to the training procedure of our teacher network and student network, which helps achieve more competitive results on the R-FOSD dataset. 

For the teacher network, we utilize the binary classifier in Section~\ref{sec:comp_discriminator} to extend training samples. Specifically, given a background image, we use the classifier to predict compatibility score for each foreground and treat those with scores larger than 0.8 as positive samples (including the ground-truth foreground), based on which  the ratio of positive and negative foregrounds per background is increased from 1:10 to 5:10. Besides, we apply data augmentation to generate additional positive and negative samples from the ground-truth foreground for each background image. Precisely, based on the ground-truth foreground, we produce 2 positive foregrounds through color jitter and Gaussian blur, and 1 negative foreground through affine transformation. After augmentation, the ratio of positive and negative samples changes from 5:10 to 7:11. 

For the student network, we select the top-$5$ foregrounds returned by  pretrained teacher network as positive samples (including the ground-truth foreground) and adopt the same negative samples as the teacher network. After applying the same data augmentation scheme as the teacher network, we also have positive and negative samples with the ratio 7:11  to train the student network. 

Finally, considering that user-provided bounding boxes may not be very accurate, we also augment bounding boxes when training teacher network and student network. Specifically, we randomly pad the bounding box with the maximum padding space being 30\% of the bounding box's width and height. 

\section{Experiments}
\label{sec:experiments}

\subsection{Dataset and Evaluation Metrics}
\label{sec:dataset_metrics}
Our S-FOSD dataset is employed for both training and testing, while R-FOSD dataset is only for testing. We employ different evaluation metrics for two datasets considering their difference in the acquisition of ground-truth foregrounds. For each metric, we report the mean evaluation results by averaging the results over all categories. Moreover, we leave the implementation details in Supplementary. 

\noindent\textbf{S-FOSD Dataset.} The training set has 57,219 pairs of foregrounds and backgrounds covering 32 categories, with a maximum of 4800 pairs and a minimum of 300 pairs in each category. The test set provides 20 backgrounds and 200 foregrounds (including 20 foregrounds from the same images as the backgrounds) for each category. The foreground/background images in the training set and test set have no overlap. Following previous works~\cite{Zhao2018CompositingAwareIS,Zhu2022GALATG}, only the foreground object from the same image is viewed as ground-truth for each background and we adopt Recall@k (R@k) as evaluation metric, which represents the percentage of background queries whose ground-truth foreground appears in top $k$ retrievals ($k=1, 5, 10, 20$ in our experiments).

\noindent\textbf{R-FOSD Dataset.} The R-FOSD dataset adopts the same foreground set as the test set of S-FOSD dataset and collects 20 backgrounds for each category. Each pair of background and foreground is shown to three human annotators to label the compatibility. The resulting dataset contains 4$\sim$190 compatible foregrounds per background, and we adopt mean Average Precision (mAP), mAP@20, and Precision@k (P@k) for evaluation, which are widely used in image retrieval and previous works~\cite{Zhao2019UnconstrainedFO,Zhu2022GALATG}. Precision@k means the percentage of compatible foreground objects in the top k retrievals, $k$=1, 5, 10, 20 in our experiments. To reduce unplausible results on the R-FOSD dataset, we exclude the objects with aspect ratios deviating more than 1.2 from that of query bounding box for all methods and compute metrics on the remaining objects.

\begin{table*}[t]
\begin{center}
\begin{tabular}{l|cccc|cccccc}
\hline
\multirow{2}*{Method} & \multicolumn{4}{c|}{S-FOSD Dataset} &\multicolumn{6}{c}{R-FOSD Dataset} \\
& R@1↑   & R@5↑  & R@10↑   &R@20↑ &mAP↑ & mAP@20↑  & P@1↑  & P@5↑ &P@10↑ & P@20↑ \\ \hline \hline
 Shape~\cite{Zhao2018CompositingAwareIS} & - & - & - & - & 50.49 & 56.22  & 47.66  & 49.72 & 50.03  &49.92  \\ 
 CFO~\cite{Zhao2018CompositingAwareIS} & 56.09 & 83.59 & 91.25 & 96.88 & 52.06 & 62.17 & 58.33 	& 56.29 & 55.90 & 52.85 \\
 UFO~\cite{Zhao2019UnconstrainedFO} & 54.69 & 81.72 & 90.94 & 95.31 & 52.73 & 63.63 & 64.12 & 59.11 & 56.30 & 53.82 \\
 FFR~\cite{Wu2021FinegrainedFR} & 57.03 & 86.25 & 93.28 & 97.97 & 53.10 & 64.12 & 61.92 & 59.64 & 57.56 & 55.06 \\
 GALA~\cite{Zhu2022GALATG} & 57.50 	& 85.17 & 93.00 & 97.33 & 52.02	& 62.83 & 62.56	& 57.33	& 55.76	& 52.93 \\
 DiscoFOS & \textbf{79.06} & \textbf{94.84} & \textbf{97.34} & \textbf{99.38} & \textbf{56.70} & \textbf{68.56} & \textbf{67.75} & \textbf{64.64} & \textbf{62.25} & \textbf{58.90} \\ \hline
\end{tabular}
\end{center}
\caption{Comparison with existing methods on our S-FOSD dataset and R-FOSD dataset. Best results are denoted in boldface.}
\label{tab:baseline}
\end{table*}

\subsection{Comparison with Existing Methods}
\label{sec:comparison}
\noindent\textbf{Baselines.} We compare our approach with previous methods~\cite{Zhao2018CompositingAwareIS, Zhao2019UnconstrainedFO, Wu2021FinegrainedFR, Zhu2022GALATG} on our two datasets. Shape~\cite{Zhao2018CompositingAwareIS} ranks foreground objects by comparing their aspect ratios with that of the query bounding box on background. CFO~\cite{Zhao2018CompositingAwareIS}, UFO~\cite{Zhao2019UnconstrainedFO}, FFR~\cite{Wu2021FinegrainedFR}, and GALA~\cite{Zhu2022GALATG} adopt the pipeline described in Section~\ref{sec:fg_bg_encoders}, which use two encoders to extract background and foreground features, and then rank the compatibility of foreground by computing its feature similarity with background. To adapt to our focus on geometric and semantic incompatibility, we replace the style feature of FFR~\cite{Wu2021FinegrainedFR} with semantic feature, in which we obtain the semantic feature by using VGG-19 \cite{Simonyan2015VGG} pretrained on ImageNet~\cite{deng2009imagenet} following~\cite{Zhao2019UnconstrainedFO}. For GALA~\cite{Zhu2022GALATG} that considers lighting and geometry compatibility, we discard lighting transformation and only keep geometry transformation~\cite{Zhu2022GALATG} to match our scenario. 
Besides, we do not compare with IFR~\cite{Li2020InterpretableFO}, as it requires pattern labels of the foreground, which are unavailable in our datasets. 
For fair comparison, we employ the same backbone network and training set for all baseline methods except Shape~\cite{Zhao2018CompositingAwareIS}. We use the same positive and negative samples for different methods when training on S-FOSD dataset.

\noindent\textbf{Quantitative comparison.} The results of baselines and our method on the proposed two datasets are summarized in Table \ref{tab:baseline}.
Moreover, when evaluating on S-FOSD dataset, the only ground-truth foreground has the same aspect ratio as the query bounding box, thus Shape~\cite{Zhao2018CompositingAwareIS} can find the ground-truth by shortcut, making the evaluation results meaningless. It can be seen that among all baselines, GALA~\cite{Zhu2022GALATG} and FFR~\cite{Wu2021FinegrainedFR} are two competitive ones, which employ contrastive learning with self-transformation~\cite{Zhu2022GALATG} and MarrNet \cite{Wu2017MarrNet3S} to enhance the perception of geometric information of foreground, respectively. Nevertheless, our method outperforms all previous approaches by an obvious margin on both S-FOSD and R-FOSD datasets.

\begin{figure}[t]
    \begin{center}
    \includegraphics[width=1\linewidth]{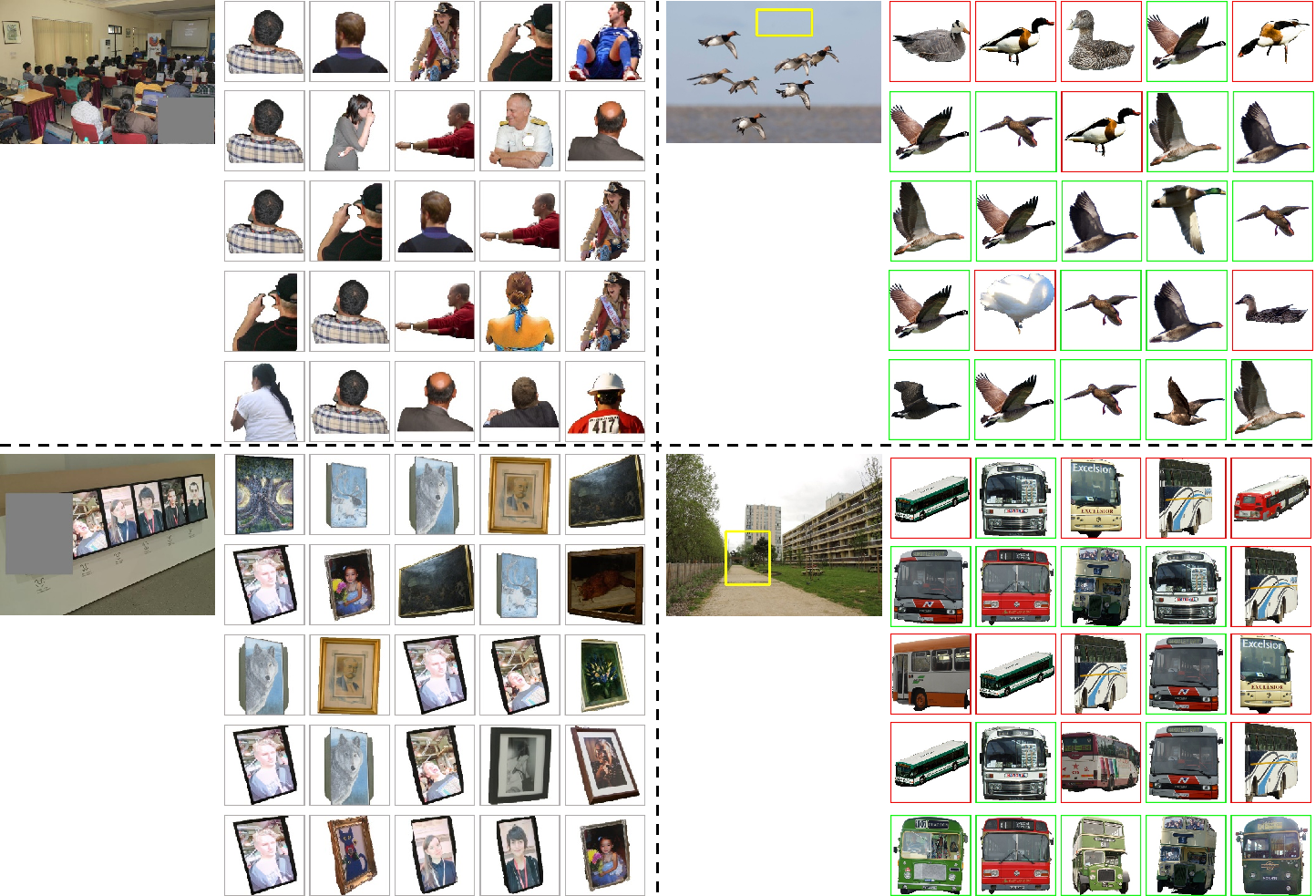}
    \end{center}
    \caption{Qualitative comparison on our S-FOSD (left) and R-FOSD (right) datasets. Each of the four examples contains a background image and several rows of the retrieval results, from top to bottom: CFO~\cite{Zhao2018CompositingAwareIS}, UFO~\cite{Zhao2019UnconstrainedFO}, GALA~\cite{Zhu2022GALATG}, FFR~\cite{Wu2021FinegrainedFR}, and ours. Additionally, in the right part, green (\emph{resp.}, red) box represents the foreground with compatible (\emph{resp.}, incompatible) label.}
    \label{fig:qualitative_results}
\end{figure}

\noindent\textbf{Qualitative comparison.} We show the retrieval results of different baselines on our two datasets in Figure~\ref{fig:qualitative_results}. For each query background, we show the returned top-5 foregrounds by different methods, which shows that our method can generally find compatible foregrounds by taking both semantic and geometric factors into account. For example, for the top-left example, given a query bounding box at the bottom-right seat, the baseline methods may return a standing or frontal person, which is unsuitable to be placed in the query bounding box. In contrast, the top returned persons by our method all have sitting posture, which appear more reasonable in terms of semantic compatibility. In the bottom-right example, given the geometry of background scene, the annotated ground-truth positive foregrounds (green box) have consistent orientation with respect to the train track in background. Our method retrieves more positive foregrounds than other approaches, demonstrating the effectiveness of our method on geometry compatibility. More qualitative results are present in Supplementary.

\subsection{Ablation Study}
\label{sec:ablation_study}

In this section, we start from the general pipeline with two encoders of previous methods \cite{Zhao2018CompositingAwareIS,Zhao2019UnconstrainedFO,Zhu2022GALATG,Li2020InterpretableFO, Wu2021FinegrainedFR} and evaluate the effectiveness of each component in our method. Different models are evaluated on our S-FOSD dataset and the results are summarized in Table~\ref{tab:ablation_study}. In row 1, we apply two encoders $\{E^b, E^f\}$ to extract background feature $\mathbf{\bar{F}}^b$ and foreground feature $\mathbf{\bar{F}}^f$, which are used to predict the background-foreground compatibility by measuring their feature similarity. It is worth mentioning that all models in Table~\ref{tab:ablation_study} rank different foregrounds by predicted compatibility scores except the row 1 and row 2, which measure compatibility by calculating feature similarity.     

\noindent\textbf{Background Encoder $E^b$.} Based on row 1 of Table \ref{tab:ablation_study}, we replace the global background feature vector $\mathbf{\bar{F}}^b$ with local background feature vector $\mathbf{\hat{F}}^b$, which is obtained by applying RoIAlign~\cite{He2020MaskR} with the crop bounding box $B$ to background feature map $\mathbf{F}^b$, resulting in worse results in row 2. This is probably because that the global background feature contains more useful information than the local background feature. So we adopt the global background feature for background encoder in other experiments.  

\noindent\textbf{Composite Image Discriminator $D$.} In our network, the discriminator plays the role of teacher. In row 9 and 10 of Table \ref{tab:ablation_study}, we feed the discriminator with the whole composite image $\mathbf{I}^c$ and the cropped composite image $\mathbf{\widetilde{I}}^c$, respectively. We see that row 10 achieves significantly better performance than row 9, which can be attributed to the aligned foreground and useful surrounding foreground context in the cropped composite. Then, the discriminator with the whole composite image is inferior to the encoders of row 1, which is roughly consistent with previous findings~\cite{Zhao2019UnconstrainedFO,Zhao2018CompositingAwareIS}. Additionally, the discriminator with cropped composite image outperforms the encoders (row 1) with a remarkable margin, which motivates us to build our method.

\begin{table}
\setlength{\tabcolsep}{1.5mm}
\begin{center}
\begin{tabular}{l|cc|cccc}
\hline
 &$E^b$  &$E^d$ & R@1↑   & R@5↑  & R@10↑   &R@20↑ \\
\hline \hline
1 &$\mathbf{\bar{F}}^b$  &        & 54.83 &	81.17 &	90.67 &	95.00      \\
2 &$\mathbf{\hat{F}}^b$  &        & 52.50 & 79.69 & 88.75 & 93.00       \\ \hline
3 &$\mathbf{\bar{F}}^b$  &$p(\cdot)_y$   & 56.72 & 87.19 & 94.69 & 97.97 \\ 
4 &$\mathbf{\bar{F}}^b$  &$[\mathbf{\bar{F}}^b, \mathbf{\bar{F}}^f]$   & 60.83 & 87.33  & 95.00 & 98.17   \\
5 &$\mathbf{\bar{F}}^b$  &$[\mathbf{\hat{F}}^b, \mathbf{\bar{F}}^f]$   & 65.16 & 90.31 & 95.31 	& 98.59  \\
6 &$\mathbf{\bar{F}}^b$  &$[\mathbf{F}^b, \mathbf{F}^f]$ & 68.28    & 90.94    & 95.63     & 98.75     \\
7 &$\mathbf{\bar{F}}^b$  &$\mathbf{\widetilde{F}}^b \oplus \mathbf{F}^f$   &64.53 	&87.66 	& 92.97  & 96.72 \\
8 &$\mathbf{\bar{F}}^b$  &$[\mathbf{\widetilde{F}}^b, \mathbf{F}^f]$   & \textbf{79.06} & \textbf{94.84}  & \textbf{97.34}  & \textbf{99.38}  \\
\hline
9 & \multicolumn{2}{c|}{$D(\mathbf{I}^c)$} & 49.22 	& 76.72  & 84.53  & 91.88    \\
10 & \multicolumn{2}{c|}{$D(\mathbf{\widetilde{I}}^c)$} & \textbf{84.38}  & \textbf{97.03}  & \textbf{98.91}  & \textbf{99.84} \\
\hline
\end{tabular}
\end{center}
\caption{The ablation studies of our method on S-FOSD dataset. $\mathbf{\bar{F}}^b, \mathbf{\hat{F}}^b$: global/local background feature vector. $\mathbf{F}^b, \mathbf{\widetilde{F}}^b$: global/local background feature map. $\mathbf{F}^f, \mathbf{\bar{F}}^f$: foreground feature map/vector.
$[\cdot, \cdot]$: feature concatenation. $\oplus$: feature map composition.
The detailed explanations can be found in Section~\ref{sec:ablation_study}. }
\label{tab:ablation_study}
\end{table}

\noindent\textbf{Knowledge Distillation Module $E^d$.} Based on row 1 of Table \ref{tab:ablation_study}, we introduce the distillation module without using feature distillation, which is essentially a compatibility classifier $p(\cdot)_y$, in row 3. The classifier takes the concatenation of foreground feature vector $\mathbf{\bar{F}}^f$ and background feature vector $\mathbf{\bar{F}}^b$ as input, and outputs the  compatibility score. The comparison between row 1 and 3 verifies that adding compatibility classification is helpful for our task.

Then we add feature distillation and explore the impact of using different forms of background-foreground interactions. In row 4 and 5, we perform interaction based on the pooled feature vectors of two encoders. Specifically, we concatenate the foreground feature vector $\mathbf{\bar{F}}^f$ with global background feature vector $\mathbf{\bar{F}}^b$ (\emph{resp.}, local background feature vector $\mathbf{\hat{F}}^b$) to feed into distillation module in row 4 (\emph{resp.}, row 5). Both results are better than row 3, proving the utility of distilling composite image feature to predict compatibility. 

Next, we perform interaction based on the last feature maps of two encoders. To this end, we explore several different ways to fuse background and foreground features. First, we directly concatenate background feature map $\mathbf{F}^b$ and foreground feature map $\mathbf{F}^f$ in row 6. By comparing row 6 and 4, it verifies the advantage of more sufficient interaction. In row 7, we extend to a more intuitive way by applying composition operation to local background feature maps $\mathbf{\widetilde{F}}^b$ and foreground feature map $\mathbf{F}^f$. Specifically, we resize the foreground feature and place it in the query bounding box on the local background feature. Unexpectedly, this leads to slight performance drop compared with row 5, which may be caused by downsampling the foreground feature. Finally, we concatenate the local background feature $\mathbf{\widetilde{F}}^b$ and foreground feature $\mathbf{F}^f$, corresponding to our complete method. As shown in row 8, our full method produces remarkably better results than the abovementioned row 6, indicating that the representation ability of the distilled feature can benefit from adequate interaction between two encoders.

\begin{table}[t]
\setlength{\tabcolsep}{1.5mm}
\begin{center}
\begin{tabular}{l|ccc}
\hline
 Model   &Time(200)↓   & Time(2000)↓ &\#Params↓ \\
\hline
Discriminator   & 700.0ms & 7000.1ms &20.03M \\
Encoders    & 4.0ms   & 4.3ms    &40.05M \\
Ours   & 5.2ms  & 6.0ms &47.13M  \\
\hline
\end{tabular}
\end{center}
\caption{Comparing inference efficiency of different models, including the discriminator $D$, encoders $\{E^b, E^f\}$, and our method in the Figure~\ref{fig:framework}. ``Time($N$)'' represents the average time of retrieving compatible foregrounds from $N$ candidates per background.}
\label{tab:time_params}
\end{table}

\subsection{Inference Efficiency Analyses}\label{sec:efficiency_analysis}
To study the efficiency of our method, we compare our method (``Ours'') with the composite image discriminator (``Discriminator''), and background and foreground encoders (``Encoders'')  on retrieval time and number of parameters, the results of which are reported in the Table~\ref{tab:time_params}. We test different models on an NVIDIA RTX 3090 GPU and measure the time of retrieving compatible foreground from $N$ candidates as retrieval time (Time($N$)). We repeat the retrieval 50 times and calculate the average time as final result.
To mimic the practical application scenarios of foreground object search, we assume that the features of all foregrounds have been extracted and saved before retrieval. In particular, for ``Encoders'' and ``Ours'', we save the feature vector and the last feature map of the foreground encoder, respectively. In this way, the retrieval time of ``Encoders'' and ``Ours'' only considers the background feature extraction and the matching between background and foreground, in which background feature extraction takes 3.1ms per image.   

We can see that the discriminator $D$ is the slowest method due to compositing with each foreground and sending each composite through forward pass.
``Ours'' runs slower than ``Encoders'' due to introducing additional interaction between background and foreground, yet only takes 6ms to retrieve from 2000 foregrounds, which enables our model for real-time applications. Regarding the number of parameters, the comparison between two encoders and ours indicates that the additional parameters (7.08M) introduced by distillation module are affordable.

\subsection{Additional Experiments in Supplementary}
\label{sec:additional_experiments}
Due to space limitation, we present some experiments in Supplementary, including quantitative comparison on different categories, generalization to new categories, the results of our method using different hyper-parameters in Eqn. (\ref{eqn:total_loss}) and Eqn. (\ref{eqn:triplet_loss}), and different ratios of positive and negative training samples, the discussion on the limitation of our method, and the results of significance test.

\section{Conclusion}
\label{sec:conclusion}
In this work, we have contributed two public datasets for Foreground Object Search (FOS), S-FOSD with Synthetic composite images and R-FOSD with Real composite images. We have also proposed a novel FOS method, which improves the general pipeline of previous methods by feature distillation. Extensive experiments on our dataset have demonstrated the utility of our proposed method on foreground object search. 

\section*{Acknowledgement}
The work was supported by the National Natural Science Foundation of China (Grant No. 62076162), the Shanghai Municipal Science and Technology Major/Key Project, China (Grant No. 2021SHZDZX0102, Grant No. 20511100300). 

{\small
\bibliographystyle{ieee_fullname}

}

\end{document}


\title{Supplementary for Foreground Object Search by Distilling Composite Feature}

\author{Bo Zhang$^{1}$ \and
Jiacheng Sui$^{2}$ \and
Li Niu\thanks{Corresponding author} $^{1}$ \and
$^1$ Center for Machine Cognitive Computing of Artificial Intelligence Institute \\ Artificial Intelligence Institute, Shanghai Jiao Tong University \\
{\tt\small\{bo-zhang, ustcnewly\}@sjtu.edu.cn} \and
$^2$ Xian Jiao Tong University \\
{\tt\small rookiecharles99@gmail.com}
}

\maketitle

In this document, we provide additional materials to supplement our main text. We will first provide more details of our Foreground Object Search (FOS) datasets and the implementation of our method in Section~\ref{sec:our_dataset} and \ref{sec:implement_details}, respectively. In Section~\ref{sec:comparison_category}, we will present the quantitative comparison between our method and baseline approaches on different categories. Meanwhile, more qualitative results of different methods will be provided in Section~\ref{sec:more_qualitative}. In Section~\ref{sec:unconstrained_retrieval}, we will demonstrate that our method can be applied to FOS for a mixture of different categories.
In Section~\ref{sec:gen_newclass}, we will apply the proposed method to new categories that have not been seen during training, which further verifies the generalization ability of our model. Then, we will study the effect of different hyper-parameters adopted in our method in Section~\ref{sec:hyper_param}, including three trade-off parameters used in our loss function and the ratio of positive and negative foregrounds per background during training stage. In Section~\ref{sec:limitation}, we will show some failure cases generated by our method and discuss the limitation of our method.

\section{Our FOS Datasets}
\label{sec:our_dataset}
Previous works~\cite{Zhao2018CompositingAwareIS,Zhao2019UnconstrainedFO,Zhu2022GALATG} on FOS did not release their datasets, which inspired us to build our own FOS datasets: S-FOSD with synthetic composite images and R-FOSD with real composite images. In this section, we will present more details about our dataset and compare our datasets with previous datasets. 

\subsection{Rules for Foreground Selection}
\label{sec:rule_fgselect}
To accommodate our task, we delete some categories and objects according to the following rules: 
1) The categories where most of the foregrounds look similar, so that most foregrounds can be considered compatible (\eg, lighthouse, apple);
2) The categories where most of the foreground objects usually appear non-independently, as parts of larger objects (\eg, clothing, wheel, flower);
3) The objects that are too large or too small in the background image (\eg, smaller than 5\% or larger than 50\% of the whole image).
4) The objects that are occluded by other objects.
5) The categories with too few remaining objects after removing occluded objects and objects with inappropriate sizes.
Summarily, the above categories and objects are either unsuitable for FOS task or beyond our focus (geometry and semantic compatibility).

\subsection{Remaining Foreground Categories}
Following the above rules, we select 32 foreground categories to construct our FOS dataset, which are airplane, bird, book, bottle, box, bread, bus, cake, camera, car, cat, coffee cup, keyboard, couch, dog, duck, fish, goose, guitar, horse, laptop, cellphone, monkey, motorcycle, pen, person, frame, taxi, toilet, train, wastebin, watch.

\begin{table*}[t]
\begin{center}
\setlength{\tabcolsep}{0.7mm}
\begin{tabular}{l|ccccccccc}
\hline
Dataset & coarse/fine & compatible factors & category & fg/category & bg & synthetic/real  & human & public \\ 
\hline \hline
CAIS-Training~\cite{Zhao2018CompositingAwareIS} & coarse    & semantics  & 8  & 2,962$\sim$38,418   & 86,800  & synthetic & N & N \\
CAIS-Evaluation~\cite{Zhao2018CompositingAwareIS} & coarse  & semantics  & 8  & 114$\sim$364   & 80  & real & Y & N \\
IFO~\cite{Li2020InterpretableFO}       & fine      & geometry, semantics    & -  & - & - & synthetic & Y & N \\
FFR-Training~\cite{Wu2021FinegrainedFR}  & fine  & geometry, style   &15   &- & 16,700 & synthetic & N  & N \\
FFR-Evaluation~\cite{Wu2021FinegrainedFR}  & fine  & geometry, style   &3   & 150 & 15 & synthetic & Y  & N \\
GALA-Pixabay~\cite{Zhu2022GALATG}     & coarse & geometry, lighting   & 914  & 912  & 833,964  & synthetic & N  & N \\  
GALA-OpenImages~\cite{Zhu2022GALATG} & coarse & geometry, lighting   & 350  & 3,926 & 1,374,344  & synthetic & N  & N \\ \hline
Our S-FOSD  & coarse    & geometry, semantics    & 32         & 500$\sim$5,000     & 57,859 & synthetic & N  & Y \\
Our R-FOSD  & coarse    & geometry, semantics    & 32         & 200                & 640    & real & Y  & Y \\ \hline
\end{tabular}
\end{center}
\caption{Comparison with previous FOS datasets. ``coarse/fine'': coarse-grained or fine-grained retrieval. ``fg/category'': foreground images per category. ``bg'': background images. ``synthetic/real'': synthetic/real composite images. ``human'': human annotation on compatibility.}
\label{tab:compare_dataset}
\end{table*}

\begin{figure*}[t]
    \begin{center}
    \includegraphics[width=1\linewidth]{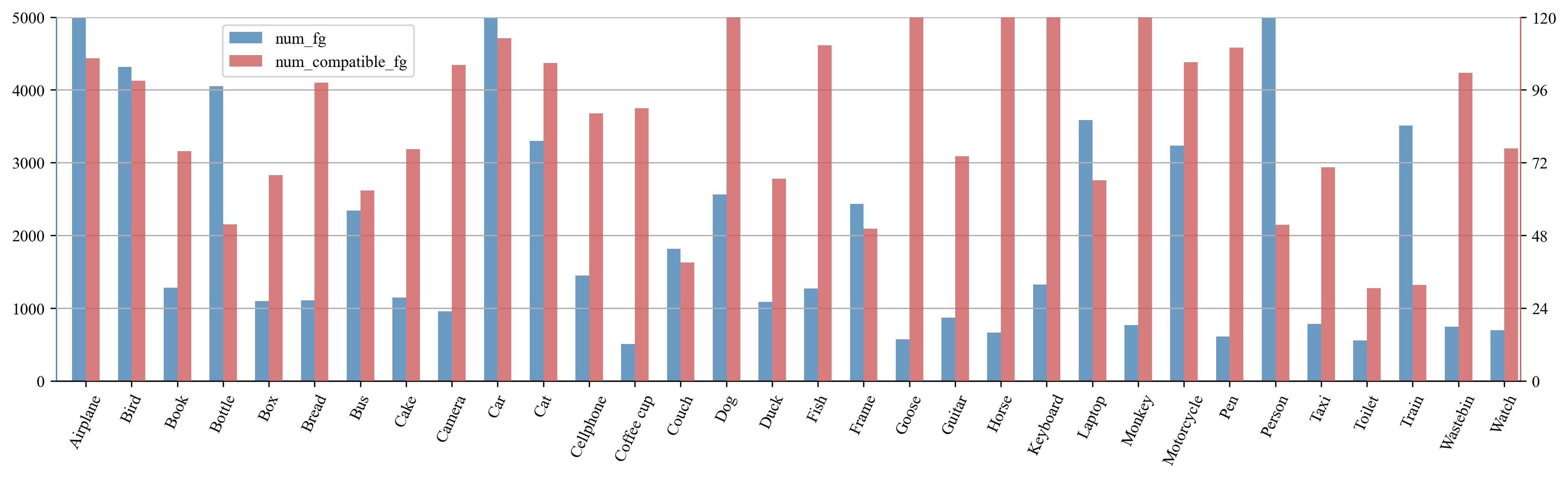}
    \end{center}
    \caption{``num\_fg'': the number of foreground images per category in our S-FOSD dataset. ``num\_compatible\_fg'': the average number of compatible foreground images per background in one category of our R-FOSD dataset, in which we provide 200 candidate foregrounds for each background and the compatibility label is assigned by three human annotators.}
    \label{fig:num_fg}
\end{figure*}

\begin{figure*}[p]
    \begin{center}
    \includegraphics[width=0.99\linewidth]{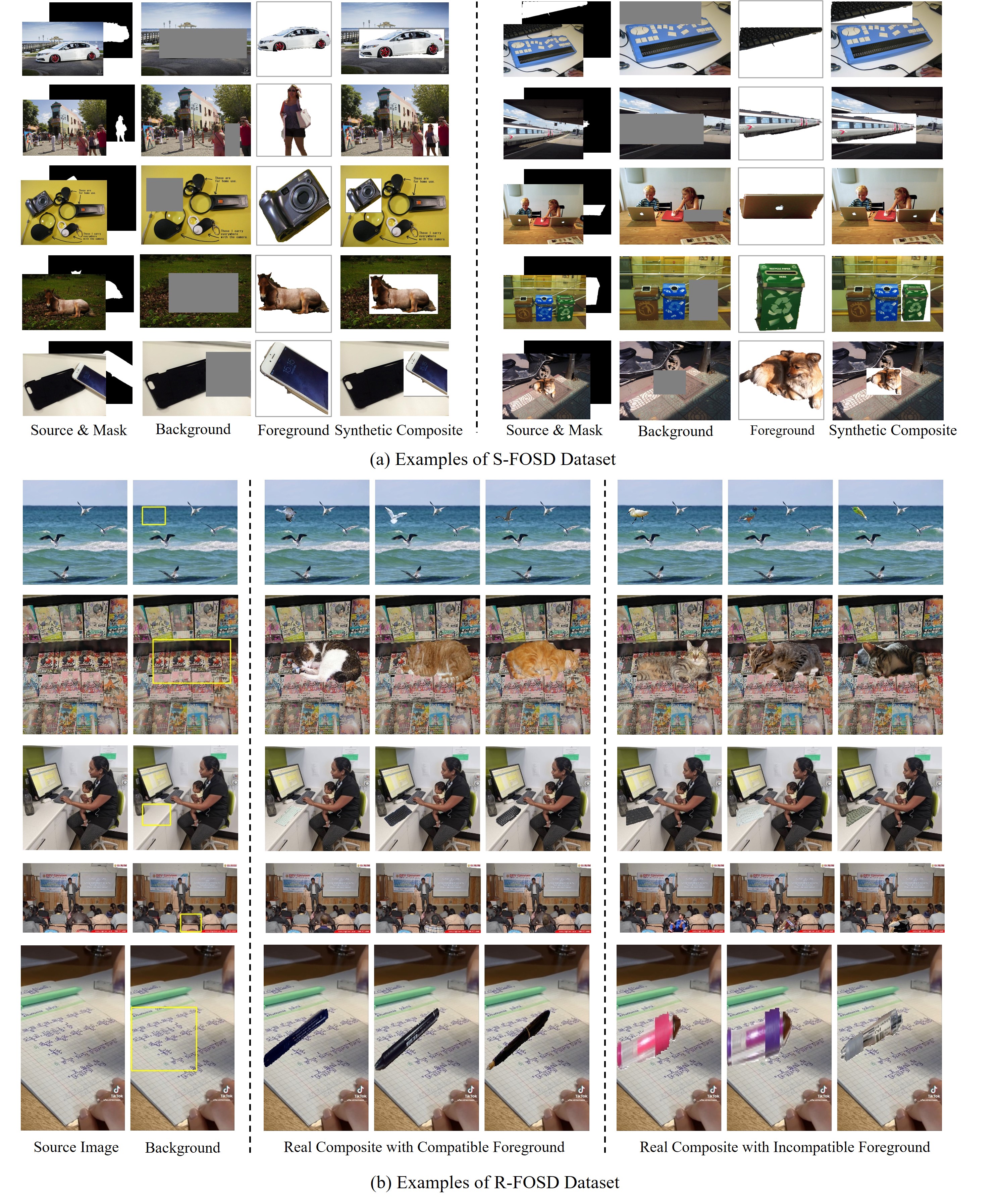}
    \end{center}
    \caption{Some examples of our S-FOSD dataset and R-FOSD dataset. For R-FOSD dataset, we show real composite images generated by placing compatible/incompatible foregrounds in the query bounding box (yellow) on background, in which compatibility label is provided by human annotators.}
    \label{fig:dataset_examples}
\end{figure*}

\subsection{S-FOSD Test Set Building}
Recall in Section 3.1 of the main text, we built the test set of S-FOSD dataset mainly concerning its diversity and quality. Here we will provide more details about the construction process. For each category, we first extract the features of all foreground images using ResNet \cite{he2016deep} pretrained on ImageNet \cite{deng2009imagenet}, and then cluster them into 100 clusters based on feature distance. Then we select the foreground objects closest to the cluster centers along with their background images as candidates and remove low-quality samples, including blurred objects, the background images whose light conditions are particularly dim, and so on. After that, we randomly select 20 background images from the remaining samples for each category. Based on the clusters where the 20 background images are located, we select 25 foreground images closest to each cluster as candidate foreground images. In this way, we obtain 20 background images and 20$\times$25 = 500 candidate foreground images for each category. After filtering low-quality images, we randomly select 200 foregrounds and 20 backgrounds per category. By selecting high-quality samples from cluster centers, we ensure the quality and diversity of test samples in S-FOSD dataset, which helps provide more effective evaluation for FOS.

\subsection{Comparison with Previous Datasets}
Following~\cite{Zhao2018CompositingAwareIS,Zhu2022GALATG}, we build our FOS datasets based on an existing large-scale real-world dataset, \ie, Open Images \cite{Kuznetsova2020TheOI}. Table~\ref{tab:compare_dataset} provides a summary comparison between our datasets and the datasets that are used in previous works~\cite{Zhao2018CompositingAwareIS, Zhao2019UnconstrainedFO, Zhu2022GALATG, Li2020InterpretableFO, Wu2021FinegrainedFR}. Among these datasets, GALA-Pixabay and GALA-OpenImages~\cite{Zhu2022GALATG} contain far more background and foreground images covering more categories, probably because that they did not filter some categories or objects like us. However, this may harm the quality of their dataset. In contrast, we remove some unsuitable categories and low-quality images to build dataset (see Section~\ref{sec:rule_fgselect}), which contributes to more effective training and evaluation on FOS task. Moreover, only the evaluation set of CAIS (CAIS-Evaluation) \cite{Zhao2018CompositingAwareIS} and our R-FOSD dataset provide real composite images, which enables more practical evaluation for real-world applications. Moreover, different from the above works, we have released our datasets to facilitate research on FOS task. 

\subsection{Dataset Statistics}
\noindent\textbf{S-FOSD Dataset.} The S-FOSD dataset contains totally 63,619 foreground images covering 32 categories. In Figure~\ref{fig:num_fg}, we show the number of foregrounds per category in S-FOSD dataset. During experiments, S-FOSD dataset is divided into training set and test set. The training set has 57,219 pairs of foregrounds and backgrounds, with a maximum of 4800 pairs and a minimum of 300 pairs in one category. The test set provides 20 backgrounds and 200 foregrounds (including 20 foregrounds from the same images as the backgrounds) for each category.

\begin{figure*}[t]
    \begin{center}
    \includegraphics[width=1\linewidth]{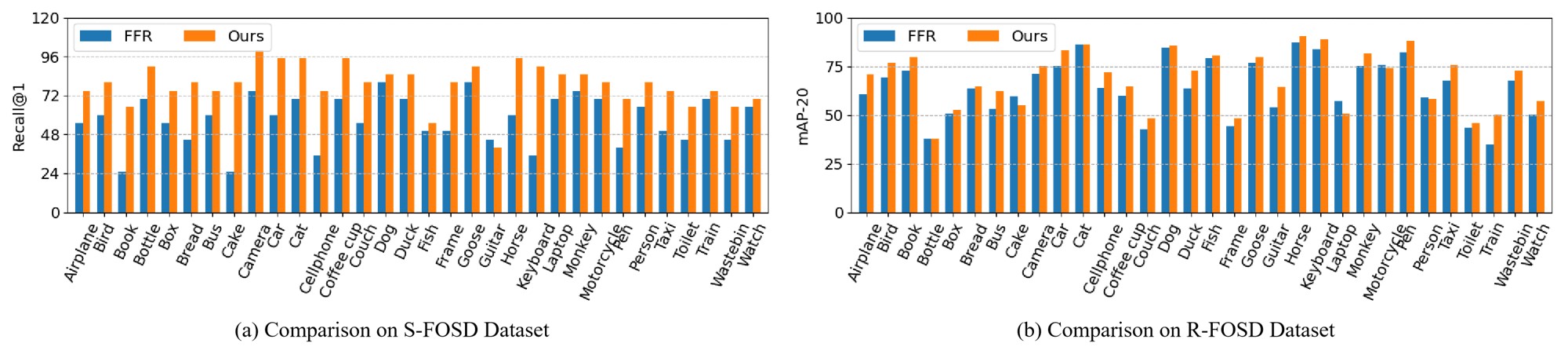}
    \end{center}
    \caption{Comparing our method with the most competitive baseline FFR~\cite{Wu2021FinegrainedFR} for each category in our S-FOSD and R-FOSD datasets.}
    \label{fig:comp_diffcls}
\end{figure*}

\noindent\textbf{R-FOSD Dataset.} The R-FOSD dataset shares the same foregrounds with the test set of S-FOSD, and has 20 backgrounds as well as 200 foregrounds per category. Each pair of background and foreground is evaluated by three human annotators in terms of their compatibility. Then, only the foreground objects that all annotators label as compatible are considered to be compatible. The resulting dataset contains 4$\sim$190 compatible foregrounds per background. We present the average number of compatible foregrounds per background in a category in Figure~\ref{fig:num_fg}.

\subsection{Visualization Examples}
In Section 3 of the main text, we have introduced the pipeline of constructing our S-FOSD and R-FOSD datasets. Here we present more examples of these two datasets in Figure~\ref{fig:dataset_examples}. Similar to Figure 2 of the main text, we show source image with instance segmentation mask, background, foreground, and synthetic composite images for one example of S-FOSD dataset. As demonstrated in Figure~\ref{fig:dataset_examples} (a), the foreground and background in S-FOSD dataset are diverse enough to cope with various real-world scenarios. For each example of R-FOSD dataset, we show source image, background, and real composite images produced by inserting foreground into the given background image. Recall there are 200 candidate foregrounds for each background. We randomly select three compatible foregrounds and three incompatible foregrounds to composite with the background, in which compatibility labels are acquired from three human annotators.   
By observing the examples in Figure~\ref{fig:dataset_examples} (b), we can roughly verify the validness of the compatibility annotations in our R-FOSD dataset.

\section{Implementation Details}
\label{sec:implement_details}
Our method is implemented using PyTorch \cite{paszke2019pytorch} and distributed on NVIDIA RTX 3090 GPU. We use the Adam optimizer~\cite{kingma2015adam} with a fixed learning rate of $1e^{-5}$ to train our model for 50 epochs. Following~\cite{Zhao2019UnconstrainedFO,Zhu2022GALATG}, we adopt VGG-19 \cite{Simonyan2015VGG} pretrained on ImageNet~\cite{deng2009imagenet} as backbone network for our discriminator $D$ and encoders $\{E^f,E^b\}$.
Before being fed into networks, both composite image and background image are directly resized to $224 \times 224$, while foreground image is first padded with white pixels to be a square image and then resized to $224 \times 224$. In this way, the composite feature map $\mathbf{F}^c$, background and foreground feature maps $\mathbf{F}^b, \mathbf{F}^f$ have the same shape $7 \times 7 \times 512$, \emph{i.e.}, $h=w=7, c=512$. We implement the knowledge distillation module $E^{d}$ with two convolution+relu operations. 
For discriminator $D$ and distillation module $E^{d}$, we append a fully-connected layer with a $sigmoid$ function as binary classifier.

Recall we generate positive and negative foregrounds from S-FOSD dataset by using a pretrained classifier (see Section 4.1 of the main text). When training on S-FOSD dataset, we adopt the same 1:10 ratio of positive and negative foregrounds per background for different models.
Additionally, we set the margin $m$ in Eqn. (2) of the main text as 0.1, and $\lambda_{kd}, \lambda_{cls}$ in Eqn. (5) of the main text as 1 via cross-validation. 

\section{Comparison on Different Categories}
\label{sec:comparison_category}
In Table 1 of the main text, we have compared with different baseline methods \cite{Zhao2018CompositingAwareIS, Zhu2015LearningAD, Zhao2019UnconstrainedFO, Wu2021FinegrainedFR} on our S-FOSD and R-FOSD datasets, in which one metric is obtained by averaging over all categories.
This comparison demonstrates that our method performs more favorably against previous baselines. To take a close look at the superiority of our method, we evaluate our model and the most competitive baseline (\ie, FFR~\cite{Wu2021FinegrainedFR}) on each single category of our S-FOSD and R-FOSD datasets, in which we report their performance in term of Recall@1 and mAP-20, respectively. As demonstrated in Figure~\ref{fig:comp_diffcls} (a), our method can generally achieve better results than FFR on different categories of S-FOSD dataset. In Figure~\ref{fig:comp_diffcls} (b), it can be seen that our model also beats FFR in most categories. These results further prove the improvement of our method over previous baseline methods on FOS task.

\begin{figure*}[p]
    \begin{center}
    \includegraphics[width=0.95\linewidth]{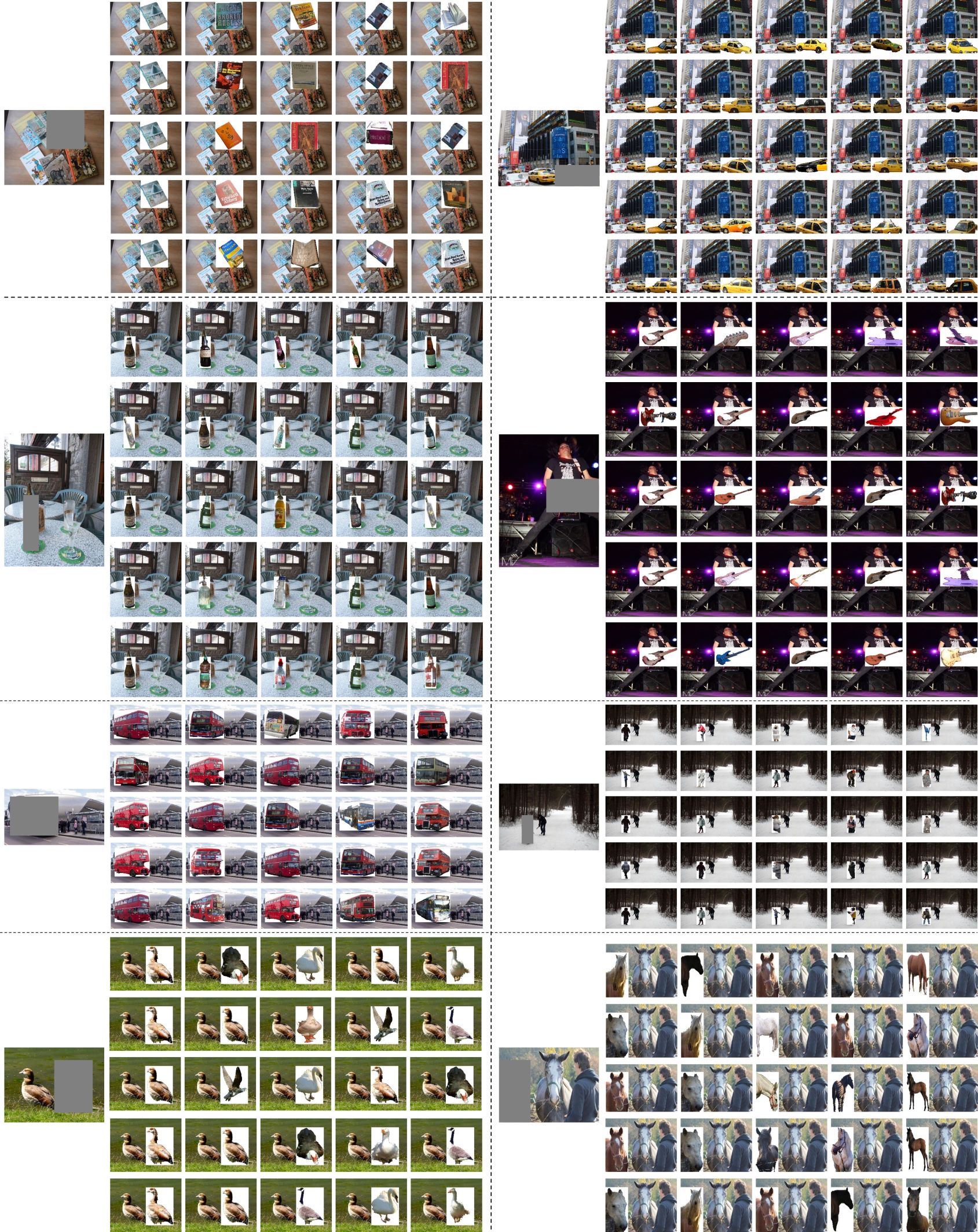}
    \end{center}
    \caption{Qualitative comparison of different methods on our S-FOSD dataset. For a background image with query bounding box, which is filled by mean image pixel, we show several rows of the retrieval results from different methods, from top to bottom: CFO~\cite{Zhao2018CompositingAwareIS}, UFO~\cite{Zhao2019UnconstrainedFO}, GALA~\cite{Zhu2022GALATG}, FFR~\cite{Wu2021FinegrainedFR}, and ours.}
    \label{fig:more_test1}
\end{figure*}

\begin{figure*}[htp]
    \begin{center}
    \includegraphics[width=0.9\linewidth]{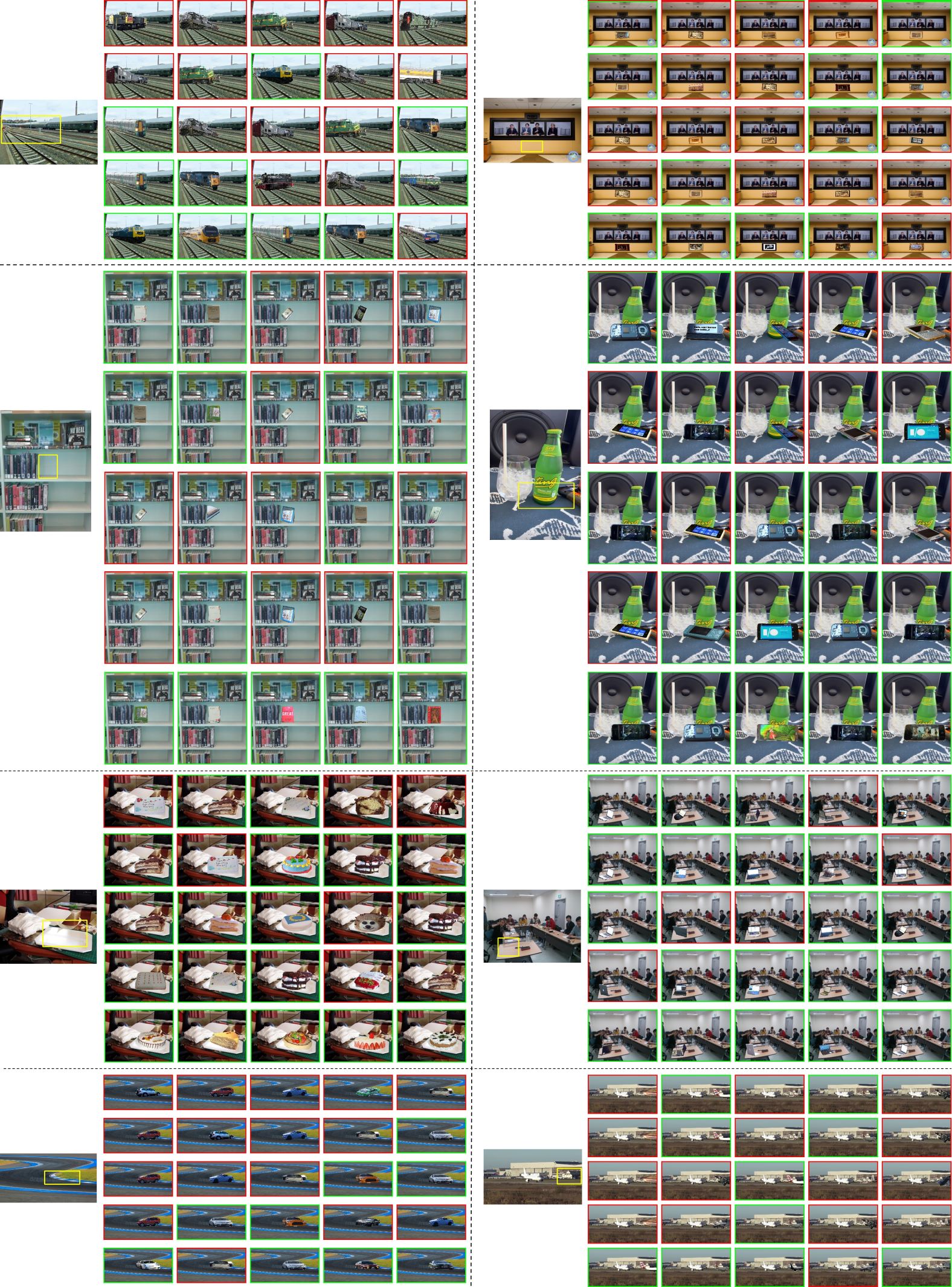}
    \end{center}
    \caption{Qualitative comparison of different methods on our R-FOSD dataset. Each example contains a background image with query bounding box (yellow) and several rows of the retrieval results from different methods, from top to bottom: CFO~\cite{Zhao2018CompositingAwareIS}, UFO~\cite{Zhao2019UnconstrainedFO}, GALA~\cite{Zhu2022GALATG}, FFR~\cite{Wu2021FinegrainedFR}, and ours. Additionally, we mark the foreground that is assigned with compatible (\emph{resp.}, incompatible) label by human annotators using green (\emph{resp.}, red) box.}
    \label{fig:more_test2}
\end{figure*}

\section{More Qualitative Results}
\label{sec:more_qualitative}
To better demonstrate the effectiveness of our method, we provide additional qualitative results of our method and baseline methods~\cite{Zhao2018CompositingAwareIS, Zhu2015LearningAD, Zhao2019UnconstrainedFO, Wu2021FinegrainedFR} on our S-FOSD and R-FOSD datasets in Figure~\ref{fig:more_test1} and Figure~\ref{fig:more_test2}, respectively. Given a background image with query bounding box, we show the composite images that are generated by placing the top-5 foregrounds of different methods in the query bounding box on the given background.
The visualized results demonstrate that our method can generally find compatible foregrounds by considering both semantic and geometric factors. For example, the background on the left of the third row in Figure~\ref{fig:more_test1} has a sloping road, indicating that inserted ``bus'' should have a matching viewpoint, so as to generate a realistic composite image. Among the compared methods, only our method works well by returning compatible foregrounds with similar viewpoint. 
In the top-right example of Figure~\ref{fig:more_test2}, our method retrieves more composite foregrounds (\ie, bird) than other baselines for the given background scene, in which a flying ``bird'' appears more suitable to be placed on the background river.
In summary, these qualitative comparisons further verify the effectiveness of the proposed method for FOS.

\begin{figure}[t]
    \begin{center}
    \includegraphics[width=1\linewidth]{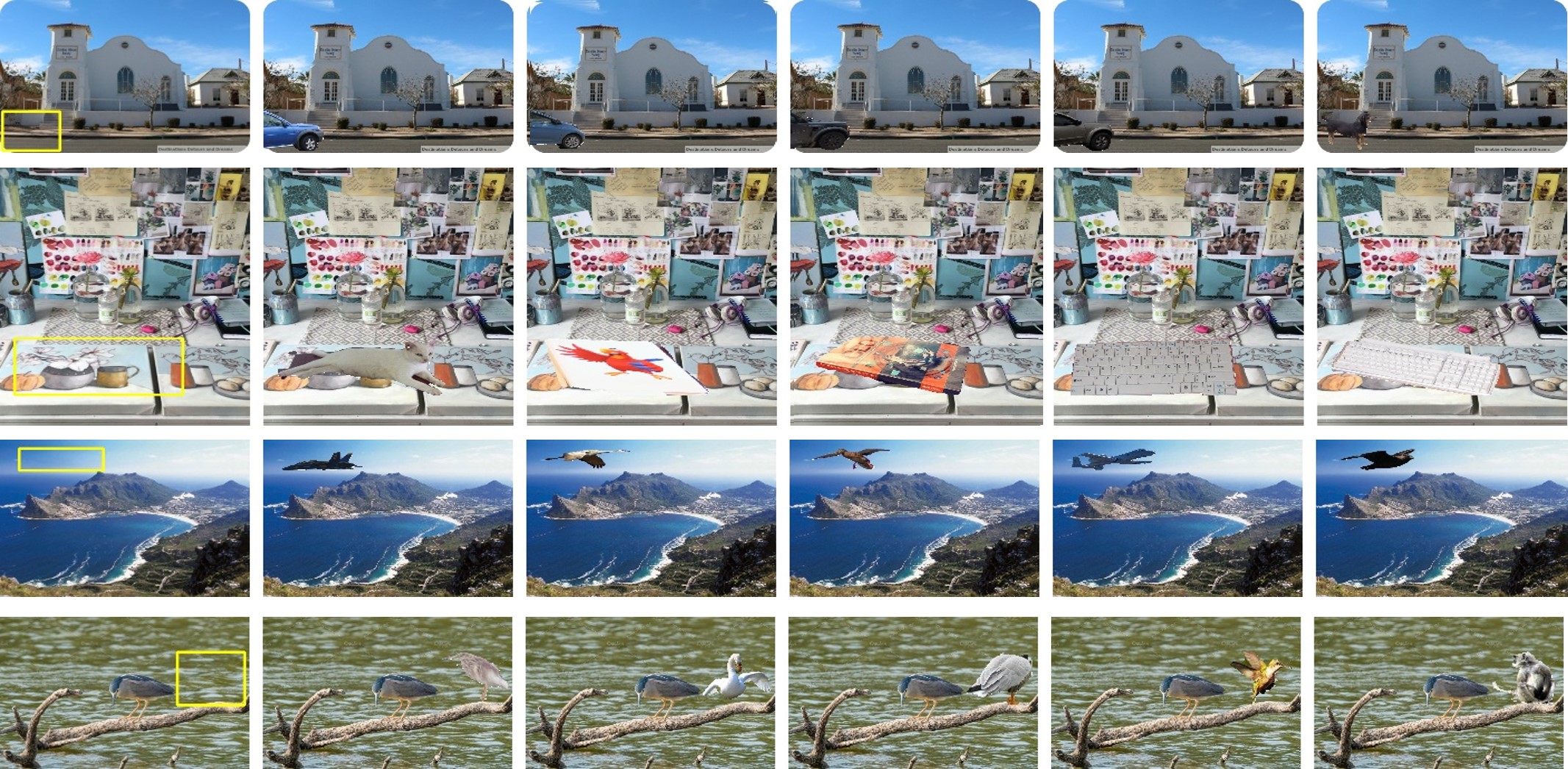}
    \end{center}
    \caption{Applying our method to FOS from 32 categories of foregrounds. In each row, our method retrieves foregrounds for the background on the left and places the retrieved foreground in the query bounding box (yellow) on background to get composite image.}
    \label{fig:unconstrained_search}
\end{figure}

\section{Retrieval from Different Categories}
\label{sec:unconstrained_retrieval}
In this work, we focus on searching compatible foregrounds from specified category for a given background, which is referred to as constrained foreground object search. In real-world application scenario, user may retrieve foreground from different categories, which is referred to as unconstrained foreground object search. To investigate the performance of our model in this scenario, we employ our model to search compatible foregrounds from 32 categories in our R-FOSD dataset. It is worth noting that the retrieval process of our method is unchanged, which means that the model ranks different foregrounds by predicted compatibility scores. We provide several examples in Figure~\ref{fig:unconstrained_search}, which demonstrate that our method is capable of generating reasonable results in this scenario as well.

\begin{figure}[t]
    \begin{center}
    \includegraphics[width=1\linewidth]{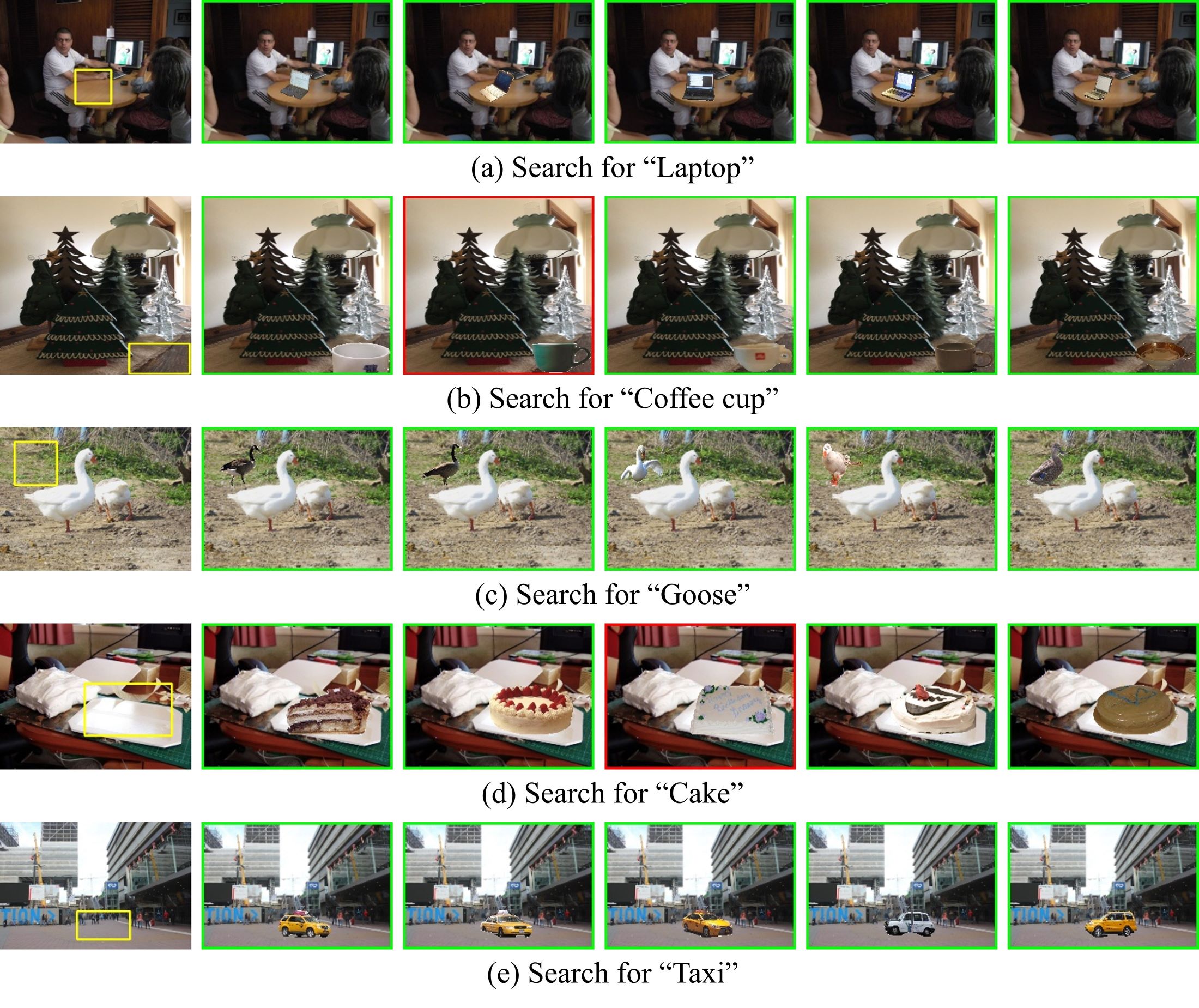}
    \end{center}
    \caption{Evaluating our method on new categories that have not been seen during training. In each row, the left is one background image with query bounding box (yellow) in R-FOSD dataset and the other are composite images with retrieved foregrounds by our method, in which compatible (\emph{resp.}, incompatible) foreground is marked with green (\emph{resp.}, red) box.}
    \label{fig:gen_newcls}
\end{figure}

\section{Generalization to New Categories}
\label{sec:gen_newclass}
To investigate the generalization ability of our learnt compatibility knowledge, we test our model on new categories that have not been seen during training. Specifically, we divide existing 32 categories of our S-FOSD dataset into five supercategories (\emph{e.g.}, animal, carrier). We then randomly choose one item from each supercategory to build the test set and the rest forms the training set. After training, we evaluate our model on our R-FOSD dataset that adopts the same foregrounds as the test set of S-FOSD dataset. 
As shown in Figure~\ref{fig:gen_newcls}, given a query background of R-FOSD dataset, our method typically can find compatible foregrounds (green box) from new categories even without training on these categories.

\begin{figure*}[htp]
    \begin{center}
    \includegraphics[width=1\linewidth]{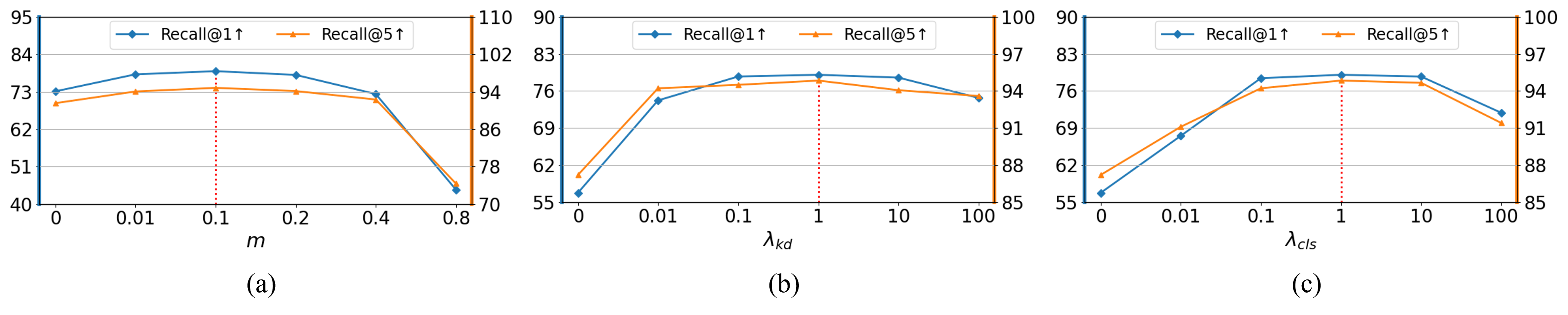}
    \end{center}
    \caption{Performance variation of our method with different hyper-parameters $m$ in Eqn. (2), $\lambda_{kd}$, $\lambda_{cls}$ in Eqn. (5) of the main text on our S-FOSD dataset. The dashed vertical lines denote the default values used in our other experiments.}
    \label{fig:hyper_param}
\end{figure*}

\section{Hyper-parameter Analyses}
\label{sec:hyper_param}
Recall that we have three hyper-parameters in the main text, \emph{i.e.}, margin $m$ for triplet loss in Eqn. (2), distillation loss weight $\lambda_{kd}$, and classification loss weight $\lambda_{cls}$ in Eqn. (5), which are respectively set as 0.1, 1, 1 via cross-validation by splitting 20\% training samples of our S-FOSD dataset as validation set. Besides, the ratio of positive and negative foregrounds per background is also considered as a hyper-parameter. In this section, we study the performance variance of our method when varying those hyper-parameters, in which we evaluate on the test set of S-FOSD dataset and report results of Recall@k (R@k) in Figure~\ref{fig:hyper_param}.

\noindent\textbf{Margin for Triplet Loss.} It is worth mentioning that our triplet loss is calculated on the cosine distance between foreground feature and background feature, which means that the margin is meaningful only if its value lies in (0, 1). To evaluate the impact of different margins $m$ for triplet loss, we vary $m$ in the range of [0, 0.8], generating results shown in Figure~\ref{fig:hyper_param} (a), in which we report the results of Recall@1 and Recall@5 on S-FOSD dataset. By comparing the results without triplet loss ($m=0$) and the results with $m=1$, we can see a clear gap between their performance, demonstrating the necessity of using triplet loss to learn discriminative foreground/background feature. When $m$ varies in the range of [0.01, 0.2], Recall@1 is in the range of [77.97, 79.06] and Recall@5 is in the range of [94.06, 94.84], which indicates that our model is robust when setting $m$ in a reasonable range.      

\noindent\textbf{Distillation Loss Weight.} With $m=0.1$, we evaluate the results of our model adopting different distillation loss weights $\lambda_{kd}$ in Figure~\ref{fig:hyper_param} (b). When $\lambda_{kd}=0$, 
the knowledge distillation module essentially degrades to a classifier and the resultant model is equivalent to the row 3 in Table 2 of the main text. In this case, the distilled feature cannot learn compatibility knowledge from composite image feature and using such feature may affect the prediction of foreground-background compatibility, leading to inferior performance. When $\lambda_{kd} \leq 1$, the performance increases as $\lambda_{kd}$ increases, which implies that adding feature distillation could benefit the compatibility prediction via distilled feature.  When $\lambda_{kd}$ becomes larger than 1, the performance begins to drop. Moreover, we find that the model can achieve satisfactory results when $\lambda_{kd}$ ranges from 0.1 to 10. 

\noindent\textbf{Classification Loss Weight.} By setting $m=0.1$ and $\lambda_{kd}=1$, we further evaluate the performance of our model with different classification loss weights $\lambda_{cls}$, the results of which are shown in Figure~\ref{fig:hyper_param} (c). For $\lambda_{cls}=0$, the predicted compatibility scores cannot be guaranteed and thus we estimate the compatibility via foreground-background feature similarity. It can be seen that the model with $\lambda_{cls}=1$ clearly outperforms the model with $\lambda_{cls}=0$, which confirms the advantages of the proposed method over the two encoders. 
Moreover, Recall@1 is in the range of [78.42, 79.06] and Recall@5 is in the range of [94.22, 94.84] when $\lambda_{cls}$ varies in the range of [0.1, 10], which implies that our model performs robust to $\lambda_{cls}$ when setting $\lambda_{cls}$ in a reasonable range. 

\begin{table}
\begin{center}
\begin{tabular}{l|cc|cccc}
\hline
  & Pos & Neg & R@1↑   & R@5↑  & R@10↑   &R@20↑ \\
\hline \hline
1 & 1 & 1 & 48.75  & 80.00  & 90.00   & 95.63  \\
2 & 1 & 5 & 77.34  & 94.00  & 96.44   & 99.38  \\ 
3 & 1 & 10 & 79.06  & 94.84  & 97.34  & 99.38  \\
4 & 1 & 20 & \textbf{80.72} & \textbf{95.23} & \textbf{98.50}  & \textbf{99.56} \\ \hline
5 & 5 & 10 & 65.63	& 87.03	& 93.59	  & 97.19 \\
6 & 10 & 10 & 46.25  & 76.09  & 86.25  & 93.59 \\ \hline
\end{tabular}
\end{center}
\caption{The performance of our method trained using different ratios of positive and negative foregrounds per background on our S-FOSD dataset. ``Pos'' and ``Neg'' indicate the numbers of positive and negative foregrounds per background, respectively.}
\label{tab:ratio_fg}
\end{table}

\noindent\textbf{Ratio of Positive and Negative Foregrounds.} Recall we generate positive and negative foregrounds from S-FOSD dataset by using a pretrained classifier in Section 4.1 of the main text. Then we set the ratio of positive to negative foregrounds per background as 1:10 for different models when training on S-FOSD dataset. To study the impact of training with different ratios, we vary the ratio and report the results of our method in Table~\ref{tab:ratio_fg}. In row 1$\sim$4, we use the only ground-truth positive foreground for one background and observe that the performance increases as the number of negative foreground increases, verifying the effectiveness of the selected negative foregrounds by the pretrained classifier. Based on row 3, we keep the number of negative foregrounds at 10 and increase the number of positive foregrounds in row 5 and 6, from which we can observe the significant performance drop. This may be attributed to the fact that the set of positive foregrounds identified by the classifier still contains some implausible samples.    
In summary, although using more training foregrounds per background may achieve better results, this would significantly slow down the training speed. To seek for the trade-off between training speed and model performance, we finally adopt the 1:10 ratio of positive and negative foreground per background for all baselines and our method in experiments.

\section{Discussion on Limitation}
\label{sec:limitation}
\begin{figure}[t]
    \begin{center}
    \includegraphics[width=1\linewidth]{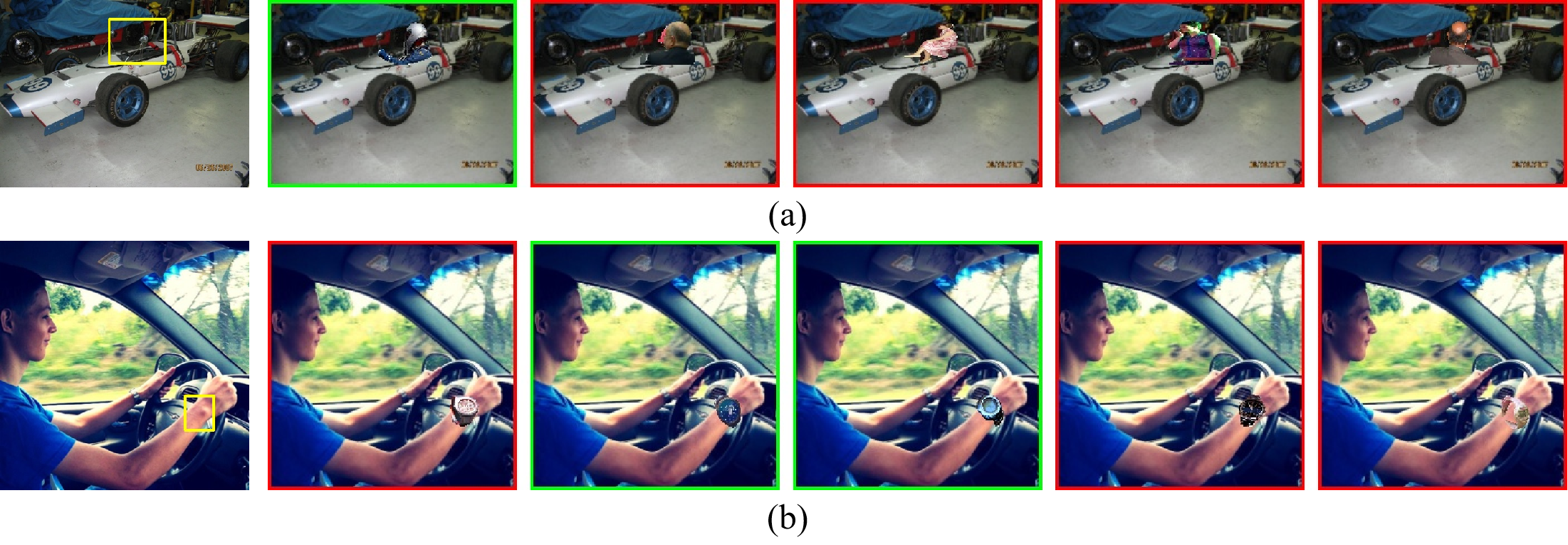}
    \end{center}
    \caption{Failure cases of our method produced on R-FOSD dataset. Given a background image with query bounding box (yellow) on the left of a row, we composite each of the returned foregrounds by our method with the given background and present them on the right, in which the compatible (\emph{resp.}, incompatible) foregrounds are indicated with green (\emph{resp.}, red) boxes.}
    \label{fig:failure_cases}
\end{figure}

Although our method is able to find compatible foregrounds for most query backgrounds, it may fail on some challenging cases. For example, as shown in Figure~\ref{fig:failure_cases} (a), all the retrieved foregrounds by our method are upper body images with non-frontal posture, yet most of them (red box) are different from the compatible one (green box) on boundary truncation and hand action, yielding implausible composite images. This can be attributed to the fact that our model mainly considers coarse-grained foreground retrieval, which makes it tougher to find foregrounds with particular attributes.         
In addition, as discussed in~\cite{Zhu2022GALATG}, the search space of FOS is bounded by the gallery of foregrounds and thus there may be a few or even no perfectly suitable foreground for a given background. In Figure~\ref{fig:failure_cases} (b), we present a such case where our method fails to return satisfactory results, because there are only a few compatible foregrounds for the given background in database.

{\small
\bibliographystyle{ieee_fullname}

}